\documentclass[review]{elsarticle}

\usepackage[british]{babel}
\usepackage{algorithm}
\usepackage{algpseudocode}
\usepackage{booktabs}
\usepackage{rotating}
\usepackage[table]{xcolor}
\biboptions{numbers,sort&compress}
\usepackage{tablefootnote}

\journal{Computers \& Industrial Engineering}

\begin{document}

\begin{frontmatter}

\title{Order Acceptance and Scheduling with Sequence-dependent Setup Times: a New Memetic Algorithm and Benchmark of the State of the Art}

\author[mymainaddress,mysecondaryaddress]{Lei He\corref{mycorrespondingauthor}}

\cortext[mycorrespondingauthor]{Corresponding author.}
\ead{helei@nudt.edu.cn}

\author[mysecondaryaddress]{Arthur Guijt}

\author[mysecondaryaddress]{Mathijs de Weerdt}

\author[mymainaddress]{Lining Xing}

\author[mysecondaryaddress]{Neil~Yorke-Smith}

\address[mymainaddress]{College of Systems Engineering, National University of Defense Technology, Changsha, China, 410073}

\address[mysecondaryaddress]{Delft University of Technology, P.O. Box
5031, 2600 GA  Delft, The Netherlands}






\begin{abstract}

The Order Acceptance and Scheduling (OAS) problem describes a class of
real-world problems such as in smart manufacturing and satellite
scheduling.  This problem consists of simultaneously
selecting a subset of orders to be processed as well as determining the
associated schedule.  A common generalization includes
sequence-dependent setup times and time windows. A novel
memetic algorithm for this problem, called Sparrow, comprises a hybridization of
biased random key genetic algorithm (BRKGA) and adaptive large
neighbourhood search (ALNS).
Sparrow integrates the exploration ability of
BRKGA and the exploitation ability of ALNS.  On a set of standard
benchmark instances, this algorithm obtains better-quality
solutions with runtimes comparable to state-of-the-art
algorithms.
To further understand the strengths and weaknesses of these algorithms,
their performance is also compared on a set of new benchmark instances with more realistic properties.
We conclude that Sparrow is distinguished by its ability to solve difficult instances from the
OAS literature, and that the hybrid steady-state genetic algorithm (HSSGA) performs well on large
instances in terms of optimality gap, although taking more time than
Sparrow.


\end{abstract}


\end{frontmatter}


\section{Introduction}

The Order Acceptance and Scheduling (OAS) problem consists of
simultaneously selecting a subset of orders to be processed as well as
determining the associated schedule.  This problem is important because
it represents a class of real-world industrial problems.  For example,
when a smart manufacturing system does not have the capacity to meet the
demand, it has to reject some orders in favour of others.
Other typical instances of the OAS problem are for example satellite
observation scheduling, where the number of observations requests is
usually higher than the number of observation a satellite can take \cite{Wang2011A},
and in industrial and commercial logistics, such as deciding the cities
to visit within a day to maximize the total profit \cite{verbeeck2017time}.

Many real-world order acceptance and scheduling problems in smart
manufacturing and other domains have setup times and time windows. The setup time is the time needed for the preparation of the next order, such as the time to prepare batches of products in the factory domain and the observation angle transition time in the satellite domain. The setup time is usually sequence-dependent, which means the setup between every two orders depends on the specific pair of orders. The time windows specify a time period for each order when it can be processed. These problems can be generalized as a typical type of OAS problem: the OAS problem with sequence-dependent setup times and time windows \cite{og2010order}. This problem has been proven to be NP-hard \cite{ghosh1997job}.

The current state-of-the-art for the OAS problem with sequence-dependent setup times and time windows limits the capability
of smart industry and operations.  Currently, realistically-sized
problems cannot be solved quickly enough: instead, solution quality must
be compromised to deliver timely solutions.  In addition, since current research focuses on improving the performance
of algorithms, few contributions try to understand the problem further
by studying how the problem properties correlate with its difficulty,
how different algorithms perform on problem instances with varying
properties, and how the instances in the standard benchmark set by Cesaret et al.~\cite{cesaret2012tabu} correspond to real-life scenarios.
These research gaps motivate our contribution.

In this article we propose a novel memetic algorithm applied for the OAS problem with sequence-dependent setup times and time windows, called Sparrow.\footnotemark The main contributions of this article are summarized as follows:

\footnotetext{This algorithm combines a population-based genetic algorithm with adaptive large neighbourhood search. Each individual in a ``swarm'' thus has a bird's eye view of the search space -- hence Sparrow.}

\begin{enumerate}
\item Sparrow is a hybridization of the biased random key genetic algorithm (BRKGA) and the adaptive large neighbourhood search algorithm (ALNS). To the best of our knowledge, this is the first hybridization of these two algorithms. We introduce several new strategies to make the hybridization efficient.

\item We compare Sparrow with state-of-the-art algorithms on a set of standard benchmark instances from the literature. The proposed algorithm obtains better-quality solutions with comparable running time.

\item We study the correlation of the problem properties and the algorithm performance and find that the congestion ratio, the length of time windows, and the correlation of processing time and revenue of orders are highly related to the difficulty of the problem.\footnotemark

\footnotetext{These terms are defined in Section~\ref{sec_property}.}

\item We further generate new instances that are more representative of real problem instances in satellite scheduling (more congestion), commerce (high correlation between revenue and processing time), and the travelling repairman problem (short processing times and long time windows), and compare the performance of multiple state-of-the-art algorithms on these new instances.
\end{enumerate}

The remainder of this article is summarized as follows: Section~\ref{sec_back} provides background information; Section~\ref{sec_alg} introduces Sparrow; Section~\ref{sec_exp} provides the empirical study of the proposed algorithm; the conclusions are summarized in Section~\ref{sec_con}.

\section{Background}\label{sec_back}

In this section we first introduce the mathematical formation of the OAS problem. Then we review the related works. Finally we describe the standard BRKGA and ALNS algorithms.

\subsection{Mathematical formulation}

Consider a set of orders $O=\{o_1,...,o_n\}$ that can be potentially scheduled. The sequence of orders is not fixed. Each order has a revenue $r_i$, a processing duration time $t_i$, a due time $d_i$, a penalty weight of tardiness $w_i$ and a time window $[b_i, e_i]$ indicating the release time and deadline of the order. Let $x_i$ be a binary decision variable representing whether order $o_i$ is selected and $p_i$ be a decision variable representing the start time of 
$o_i$. Let $T_i$ be the tardiness of $o_i$, $T_i=\max\{p_i+t_i-d_i,0\}$. The problem can be formulated as a mixed integer programming (MIP) model:
\begin{equation} \label{obj}
\textrm{Maximize} \quad\sum_{i=1}^n x_i(r_i-w_iT_i)
\end{equation}
subject to
\begin{equation} \label{con1}
\begin{array}{c}
\max\{b_j,p_i+t_i\}+s_{ij}\le p_j \\
\forall i,j\in \{1,...,n\} \quad\textrm{if}\quad x_i=1,x_j=1,p_i<p_j
\end{array}
\end{equation}
\begin{equation} \label{con2_1}
T_i=\max\{p_i+t_i-d_i,0\}\quad \forall i\in \{1,...,n\}
\end{equation}
\begin{equation} \label{con2}
b_i\le p_i \le e_i-t_i \quad \forall i\in \{1,...,n\} \quad\textrm{if}\quad x_i=1
\end{equation}
\begin{equation} \label{con3}
x_i \in \{0,1\} \quad \forall i\in \{1,...,n\}
\end{equation}

The objective function (\ref{obj}) maximizes the total reward of scheduled orders. The reward of an order equals its revenue minus the penalty of tardiness. Constraints (\ref{con1}) ensure the time between every two orders should be long enough for the setup, where $s_{ij}$ is the setup time between orders $o_i$ and $o_j$. The value of $s_{ij}$ depends on $i$ and $j$ and is always non-negative. Constraints (\ref{con2_1}) define the domains of the tardiness of orders. Constraints (\ref{con2}) and (\ref{con3}) define the domains of the decision variables $p_i$ and $x_i$ respectively.

A feasible solution of this problem is a sequence of orders with scheduled start times which meet all the constraints above. It can be represented as a tuple $(X,P)$, where $X=\{x_1,...,x_n\}$ and $P=\{p_1,...,p_n\}$.

\subsection{Related works}

Due to the large number of OAS variants, we only review the articles studying the OAS problem with sequence-dependent setup times and time windows. Readers are referred to Slotnick~\cite{slotnick2011order} for a comprehensive survey of the OAS problem.

The OAS problem with sequence-dependent setup times and time windows was first proposed by O{\u{g}}uz et al.~\cite{og2010order}. They proposed a mixed integer linear programming (MILP) method for this problem, two simple greedy constructive heuristics, and a heuristic algorithm called iterative sequence first-accept next (ISFAN), which is based on the simulated annealing (SA) algorithm. Cesaret et al.~\cite{cesaret2012tabu} proposed a tabu search (TS) algorithm to solve this problem, which achieved better results than ISFAN. Another major contribution of this article is that it provided the benchmark set and the upper bounds of this problem. Nguyen et al.~\cite{nguyen2014enhancing} proposed an exact branch-and-bound (B\&B) with genetic programming (GP) to discover good ordering rules. Their method could find optimal solutions for instances with up to 20 orders. Silva et al.~\cite{silva2018exact} proposed a new arc-time-indexed formulation and two exact methods based on Lagrangian relaxation and column generation respectively. They computed tighter upper bounds compared with those by Cesaret et al.~\cite{cesaret2012tabu}. They also proposed an iterated local search (ILS) method to solve the problem.

The hybridization of population-based methods with local search (LS) has been used a lot to solve this problem. Lin and Ying~\cite{lin2013increasing} proposed an artificial bee colony (ABC) algorithm to solve this problem. They used a permutation based representation and generated new solutions through standard order crossover or a LS algorithm (removing and inserting orders). Their methods achieved better results than those of TS \cite{cesaret2012tabu}. Chen et al.~\cite{chen2014diversity} proposed a diversity controlling genetic algorithm (DCGA), which selects individual chromosomes not only depending on the fitness, but also on the diversity of the whole population. In their algorithm, they also adopted similar LS methods as in \cite{lin2013increasing}. Nguyen et al.~\cite{nguyen2015dispatching} proposed a dispatching-rule-based genetic algorithm (DRGA). The dispatching rule is a strategy to calculate some heuristic priorities for orders and these priorities are used in the decoding process. They also used a simple greedy LS method to improve the solutions. Following Nguyen et al., Park et al.~\cite{park2013evolving} proposed a GP method with stochastic dispatching rules, which could generate and evolve rules to find good solutions. They later proposed a hybrid particle swarm optimization (PSO) method with TS using dispatching rules learned by GP \cite{park2014enhancing}. Similar ideas can also be found in \cite{nguyen2014sequential}. Nguyen~\cite{nguyen2016learning} proposed a hyper-heuristic algorithm, which is a learning and optimizing system (LOS). The method trains a set of optimizing rules in the learning phase, then the rules are used by a genetic algorithm (GA) to find good solutions in the optimizing phase. More recently, Chaurasia and Singh~\cite{chaurasia2017hybrid} proposed a hybrid steady-state genetic algorithm (HSSGA) and an evolutionary algorithm with guided mutation (EA/G-LS). The authors also proposed an ABC-based hyper-heuristic. However they only published the results on small instances \cite{chaurasia2019artificial}.

From the above review we can find that the hybridization of population-based methods with LS methods is a promising method for the OAS problem with sequence-dependent setup times and time windows.  The evolutionary algorithm (EA) is an important population-based algorithm. The hybridization of EA and LS is called memetic algorithm (MA) \cite{moscato1989evolution} and has also been successfully applied to problems where all orders are scheduled including the travelling salesman problem \cite{bontoux2010memetic}, the flowshop scheduling problem \cite{wang2017nsga} and the location routing problem \cite{asgari2017memetic}. It is recognized that the superior performance of MA comes from the integration of the exploration ability of the EA and the exploitation ability of the LS \cite{krasnogor2005tutorial}. A detailed description and comprehensive survey of MA could be found in \cite{neri2012memetic}.

Next, we review the two components of Sparrow: BRKGA and ALNS.

The BRKGA algorithm is a variant of the EA method proposed by Gon{\c{c}}alves and De~Almeida~\cite{gonccalves2002hybrid}. Compared with standard GA methods, BRKGA offers more flexibility in encoding solutions \cite{chaves2018adaptive} and produces as good or better solutions \cite{gonccalves2011biased}. BRKGA has shown competitive performance on a series of optimization problems \cite{prasetyo2015survey}, including the satellite scheduling problem \cite{tangpattanakul2015biased}, which is very similar to the problem studied in this article.  The ALNS algorithm was first proposed by Pisinger and Ropke~\cite{pisinger2007general}. It performs particularly well on problems with order acceptance features, such as the orienteering problem \cite{palomo2017planning}, the satellite scheduling problem \cite{liu2017adaptive} and the pickup and delivery problem with selective requests \cite{li2016adaptive}.

Comparing the two approaches, the advantage of BRKGA is that it can search the large solution space efficiently. However, the problem-independent nature of BRKGA makes it difficult to find high-quality solutions. For the second approach, ALNS, multiple neighbourhood operators can be defined according to the characteristics of the problem. Therefore ALNS has a good exploitation ability and can adapt itself according the different properties of problem instances. However, its search efficiency
can founder due to the entrapment in a local optimum because of the single-point search. We are interested to see if a hybridization could be made such that the advantages of the two algorithms can be integrated and achieve better performances.

\subsection{The standard BRKGA and ALNS algorithms}

In this section we introduce the standard BRKGA \cite{gonccalves2002hybrid} and ALNS \cite{pisinger2007general}.

In BRKGA, a population of $p$ solutions are represented as $p$
chromosomes, consisting of genes which are encoded by real values randomly generated in the interval [0,1], i.e., random keys. The number of genes in a chromosome equals the number of orders. Chromosomes are then decoded to obtain solutions and calculate fitness, which is a problem-dependent process. Example 1 shows a simple decoding method for the problem studied in this article. The best $p_e$ chromosomes are recognized as elite individuals. In the mutation operation, $p_m$ chromosomes are generated randomly (i.e., same as the initial population) to avoid the entrapment in a local optimum. In the crossover operation, $p-p_e-p_m$ chromosomes are generated by inheriting each gene from an elite parent with the probability $\rho_e$ and a non-elite parent with the probability $1-\rho_e$. The encoding strategy of BRKGA ensures that there are no duplicate orders and all solutions are feasible in the crossover operation.

\vspace{2mm}
\textit{Example 1.} All the orders are first sorted according to the ascending gene values. Then the orders are started as early as possible. Any order that can not complete before the deadline or can not achieve a positive reward, will be rejected. Consider a set of five orders $\{1, 2, 3, 4, 5\}$ with the corresponding gene values $\{0.6, 0.2, 0.3, 0.5, 0.7\}$. The decoder tries to start each order as early as possible, following the sequence of $\{2, 3, 4, 1, 5\}$. Assume that the end time of order 3 is larger than the deadline of order 4, which means that order 4 cannot be inserted in the solution, the resulting solution would be $\{2, 3, 1, 5\}$.
\vspace{2mm}

ALNS starts from an initial solution usually generated by a simple heuristic, because it is less sensitive to the initial solution than general local search~\cite{demir2012adaptive}. ALNS proceeds to generate new solutions through destroying and repairing.  In the destroying process, $p_d$ orders are removed from the current solution by removal operators.  The unscheduled and removed orders are then inserted into the destroyed solution in the repairing process by insertion operators.  There are multiple removal and insertion operators.  At each iteration, a pair of removal and insertion operators is selected by a roulette wheel mechanism according to their weights. After a certain number of iterations, the weight of the operator $w_i$ is updated adaptively according to its accumulated score $\pi_i$ in the previous iterations, $w_{i}=(1-\lambda )w_{i}+\lambda {\pi _{i}}/\sum_j\pi _{j}$, where $\lambda \in [0,1]$ is a reaction factor which controls how sensitive the weights are to changes in the performance of operators. A simulated annealing (SA) criterion is used to control the acceptance of new solutions by a temperature parameter $T$. Let $f(S)$ and $f(S')$ be the reward of current solution $S$ and new solution $S'$ respectively. The new solution $S'$ is accepted if $f(S') > f(S)$; otherwise, it is accepted with probability: $\rho =\exp \left( \frac{100}{T}\left( \frac{f(S')-f(S)}{f(S)} \right) \right)$.

\section{Sparrow}\label{sec_alg}

In this section, we first introduce the main framework of Sparrow. The tight hybridization aims to avoid the possible high running time of the combination of the two complex algorithms and aims to integrate the advantages of them.
Then we introduce the BRKGA part, where we propose a new bounded-width gene encoding strategy for the problem with time windows, a hybrid decoding method for the OAS problem and an intelligent crossover operator. Finally we introduce the ALNS part, where we define several neighbourhood operators and propose a fast insertion algorithm considering sequence-dependent setup times.

\subsection{The main framework}

\begin{figure}[htbp]
\centering
\includegraphics[height=1.58in]{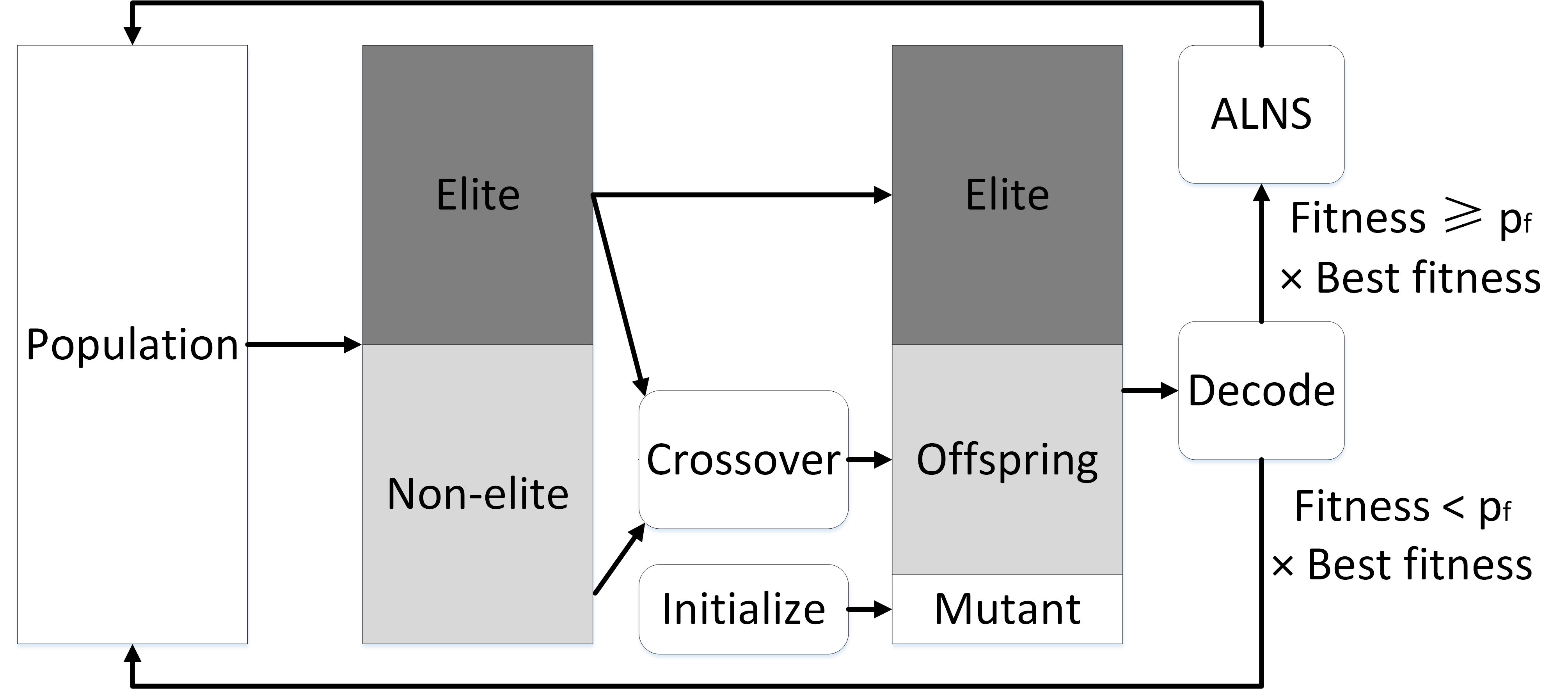}
\caption{The main framework of Sparrow}\label{figsparrow}
\end{figure}

The main framework of Sparrow is shown in Algorithm~\ref{alg1}. The main structure of this algorithm is derived from BRKGA. ALNS is used in each iteration to improve the quality of the population. Figure~\ref{figsparrow} shows how the algorithm evolves generations of solutions.

A problem of integrating ALNS in BRKGA is the complexity, because ALNS itself needs multiple iterations to improve a single solution. In order to hybridize these two algorithms without increasing the computation time too much, we run
the destroying and repairing process of ALNS only once for each solution in the population and only on relatively good solutions with at least $p_f$ of the best fitness (i.e., the value of the objective function Eq.~(\ref{obj})) found so far (Algorithm~\ref{alg1}, Lines 6--8).

In standard ALNS, the algorithm updates the temperature in each iteration and updates the weights of operators after a certain number of iterations according to their performances. In this population-based ALNS, these two processes happen in each iteration (Algorithm~\ref{alg1}, Lines 19, 20). The weights of operators are updated according to their performances in improving the individuals in the population.

There are three terminal conditions of Sparrow: when the maximum iteration is reached; when the best fitness is not improved for a maximum number of consecutive iterations; when all orders in the instance are scheduled and receive full revenue.

\begin{algorithm}[tb]
\footnotesize
\caption{Sparrow}\label{alg1}
\begin{algorithmic}[1]
\State Generate an initial population $P$, consisting of $p$ chromosomes;
\State Let $S^*$ be the best solution and $f^*$ be its fitness, $S^*\leftarrow \emptyset$, $f^*\leftarrow 0$;
\Repeat
\For{each chromosome $c$ in $P$}
\State $S\leftarrow Decode(c)$
\If{$f(S) \ge p_{f}f^* $}
\State $S\leftarrow ALNS(S,f^*)$
\EndIf
\If{$f(S)>f^*$}
\State $f^*\leftarrow f(S)$, $S^*\leftarrow S$
\EndIf
\EndFor
\State Sort chromosomes in $P$ according to a descending order of fitness;
\State Add the top $p_e$ chromosomes to Elite set $E$;
\State Add the remaining chromosomes to Non-elite set $N$;
\State Generate a mutation set of $p_m$ chromosomes: $M\leftarrow Mutation()$;
\State Generate a crossover set of $p-p_e-p_m$ chromosomes: $C\leftarrow Crossover(E,N)$;
\State $P\leftarrow E\cup M \cup C$;
\State Update the temperature of ALNS with the coefficient of annealing $c_a$: $T\leftarrow c_aT$;
\State Update the weights of different operators of ALNS;
\Until Terminal condition is met;
\State \Return $S^*$, $f^*$;
\end{algorithmic}
\end{algorithm}

\subsection{The BRKGA}

\subsubsection{Encoding and decoding}

In standard BRKGA, the initial random key is generated randomly in the interval [0,1] and then decoded according to the simple decoding method in Example 1. However, for this problem with time windows, if an order with early time windows receives a large random key, this order may not be scheduled because its time window is occupied by other orders with smaller random keys. To solve this problem, we propose the bounded-width gene encoding method and the hybrid decoding method.

\textbf{Bounded-width gene encoding.} When assigning the initial gene value to orders, we calculate the proportion of the window size in the whole scheduling horizon. For example, if an order has the time window [0,10] and the whole scheduling horizon is [0,100], the interval for the gene value of this order is [0,0.1]. This strategy helps to reduce the solution space, especially for the instances with short time windows.

One problem of this strategy is that orders with earlier time windows are preferred than those with later time windows, because those with later time windows receive larger random keys and may not be scheduled because of the early ones. This problem is solved by the ALNS algorithm in the removal process. The details are shown in Section~\ref{sec_operator}.

\textbf{Hybrid decoding method.} Our hybrid decoding method consists of the simple decoding method as in Example 1 and a complex decoding method with order insertion. For the complex decoding, all the orders are also sorted according to the ascending gene values. Instead of starting each order following this sequence, this decoder tries to insert each order into the current partial solution at the position which increases minimum setup time. When inserting an order, the orders in the partial solution might be postponed but will not be canceled. The detailed insertion strategy is introduced in Section~\ref{sec_insert}.

The complexity of the complex decoding is certainly higher than the simple decoding. However, it helps to increase the probability of orders with short time windows being successfully scheduled. We use an adaptive strategy to hybridize these two strategies according to the instance. We calculate the value of the average length of all orders divided by the scheduling horizon. Let this value be $r_d$. For each chromosome, the algorithm chooses the simple decoding with the probability of $r_d$; otherwise it uses the complex decoding.

After a number of iterations, the random keys of different orders may squeeze together due to the following crossover and ALNS operations. In order to avoid this, we normalize the keys to distribute them evenly in the interval of [0,1] in each iteration.

\subsubsection{Intelligent crossover}

In the standard BRKGA, the offspring inherits each gene from the elite parent with the probability $\rho_e$; otherwise it inherits the gene from the non-elite parent. This strategy ensures that the offspring has more genes from the elite parent. However, in this OAS problem with sequence-dependent setup times, the quality of a sequence of orders is important. The setup time between any two orders can differ much. The standard crossover operation might break good sequences when it inherits genes from different parents, thus producing low-quality offspring.

We propose an intelligent crossover strategy to keep good order pairs together in the offspring. Before the crossover, the algorithm identifies good order pairs in the current elite and non-elite parents. A pair of orders is recognized as $good$ if it has a relatively high unit reward (i.e., the total reward of the two orders divided by the time between the start time of the preceding order and the end time of the following order is higher than a constant parameter $f_g$).

When a good pair of orders is found, the gene value of the following order will be changed to the gene value of the preceding order plus a small enough positive number $\epsilon$, to ensure they will have a high probability of being together in the decoding phase. The crossover operator selects genes from one of the parent one-by-one. Normally there is a constant $\rho_e$ determining the probability whether the gene comes from the elite or the non-elite parent. For the intelligent crossover, when one gene of a good pair is selected, the probability that the other gene in the good pair will be selected is enhanced to be $p_g$, a value close to 1.

The detailed process is shown in Algorithm~\ref{algcross}.

\begin{algorithm}[htbp]
\footnotesize
\caption{Intelligent crossover}\label{algcross}
\begin{algorithmic}[1]
\State \textbf{Input:} Elite set $E$; Non-elite set $N$;
\State Let $C \leftarrow \emptyset$ be the crossover offspring set;
\Repeat
\State Select an elite parent chromosome $C_e$ from $E$ randomly;
\State Select a non-elite parent chromosome $C_n$ from $N$ randomly;
\State Initialize a child chromosome $C_c$ with an empty gene list;
\State Identify and label $good$ pairs of orders in $C_e$ and $C_n$;
\For{$i \leftarrow 1,i\le n,i$++}
\If{the $i^{th}$ gene in $C_e$ is $good$ $\wedge$ its pair is in $C_c$}
\State Generate a random value $r$ in [0,1];
\If{$r < p_g$}
\State Add the $i^{th}$ gene in $C_e$ into $C_c$;
\Else
\State Add the $i^{th}$ gene in $C_n$ into $C_c$;
\EndIf
\Else
\If{the $i^{th}$ gene in $C_n$ is $good$ $\wedge$ its pair is in $C_c$}
\State Generate a random value $r$ in [0,1];
\If{$r < p_g$}
\State Add the $i^{th}$ gene in $C_n$ into $C_c$;
\Else
\State Add the $i^{th}$ gene in $C_e$ into $C_c$;
\EndIf
\Else
\State Generate a random value $r$ in [0,1];
\If{$r < \rho_e$}
\State Add the $i^{th}$ gene in $C_e$ into $C_c$;
\Else
\State Add the $i^{th}$ gene in $C_n$ into $C_c$;
\EndIf
\EndIf
\EndIf
\EndFor
\Until $|C|=p-p_e-p_m$
\State \textbf{return} $C$;
\end{algorithmic}
\end{algorithm}

\subsection{The ALNS}

As mentioned above, if a solution has a relatively good fitness, ALNS is used to improve the solution further. In the following sections, we first introduce the neighbourhood operators; then we introduce a fast insertion algorithm used in the insertion operator to insert orders into the current solution. The full ALNS algorithm is shown in Algorithm~\ref{algalns}. In Algorithm~\ref{algalns}, $\sigma_1>\sigma_2>\sigma_3$ are three score increment parameters, used to increase the scores of different neighbourhood operators depending on their performances.

\begin{algorithm}[htbp]
\footnotesize
\caption{The ALNS algorithm}\label{algalns}
\begin{algorithmic}[1]
\State \textbf{Input:} Current solution $S_C$; best fitness $f^*$;
\State Add all unscheduled orders in order bank $B$ according to $S_C$;
\State Select a removal operator $O_R$ according to the weights;
\State Sort the scheduled orders in $S_C$ according to $O_R$;
\State $S_N \leftarrow$ Remove the top $p_d$ orders from $S_C$ and add them into $B$;
\State Update the start times and time slacks of orders in $S_N$;
\State Select an insertion operator $O_I$ according to the weights;
\State Sort the unscheduled orders in $B$ according to $O_I$;
\For{each candidate order $o_c$ in $B$}
\State $S_N\leftarrow$ $FastInsertionAlgorithm(S_N,o_c)$
\EndFor
\If{$f(S_N)>f^*$}
\State $\pi_{O_R}\leftarrow \pi_{O_R}+\sigma_1$, $\pi_{O_I}\leftarrow \pi_{O_I}+\sigma_1$, $S_C\leftarrow S_N$;
\Else
\If{$f(S_N)>f(S_C)$}
\State $\pi_{O_R}\leftarrow \pi_{O_R}+\sigma_2$, $\pi_{O_I}\leftarrow \pi_{O_I}+\sigma_2$, $S_C\leftarrow S_N$;
\Else
\If{The SA criterion accepts $S_N$}
\State $\pi_{O_R}\leftarrow \pi_{O_R}+\sigma_3$, $\pi_{O_I}\leftarrow \pi_{O_I}+\sigma_3$, $S_C\leftarrow S_N$;
\EndIf
\EndIf
\EndIf
\State \textbf{return} $S_C$;
\end{algorithmic}
\end{algorithm}

\subsubsection{Neighbourhood operators}\label{sec_operator}

In order to ensure ALNS is suitable for a diverse range of problem instances with varying characteristics, we use the following five removal operators and two insertion operators. These operators are adapted from those in the literature \cite{pisinger2007general,demir2012adaptive} to fit our problem.

The five removal operators are: random ($p_d$ orders are removed randomly); min revenue ($p_d$ orders with lower revenue are removed); min unit revenue ($p_d$ orders with lower unit revenue are removed: the unit revenue is the revenue of the order divided by its processing time); max setup time ($p_d$ orders with longer setup time are removed); sequence removal (the worst sequence of $p_d$ orders is removed: the quality of a sequence is evaluated by the unit revenue).

The two insertion operators are: max revenue; max unit revenue. The insertion operator first sorts the orders according to the descending revenue or unit revenue, then tries to insert orders in the current solution using the fast insertion algorithm in Section~\ref{sec_insert}.

In order to make sure that the solution operated by ALNS can be encoded correctly for the following BRKGA operations, when removing an order, the gene of this order will be assigned a random real value larger than the largest gene in the current solution and smaller than 1; when inserting an order, the gene of this order will be assigned a random real value between the gene values of its preceding and following orders. This strategy helps to solve the problem brought by the bounded-width gene generation strategy. Although the orders with earlier time windows have smaller gene values in the initialization, they can be removed and get a large gene value in ALNS.

\subsubsection{Fast insertion algorithm}\label{sec_insert}

Our last innovation is a fast insertion algorithm, which first evaluates the feasibility and the cost of all the positions rapidly by a concept called \textit{time slack}. Then the best position is selected to insert the order. The insertion algorithm is used in the repairing process when we insert orders back to the solution. The detailed process of the fast insertion algorithm is shown in Algorithm~\ref{alginsert}.

\begin{algorithm}[htbp]
\footnotesize
\caption{Fast insertion algorithm}\label{alginsert}
\begin{algorithmic}[1]
\State \textbf{Input:} Current solution $S_C$, candidate order $o_c$;
\For{each scheduled order $o_i$ in $S_C$}
\If{$e_c> b_i$}
\State $t_1\leftarrow$ Calculate the end time of $o_c$ if it is inserted after $o_i$;
\State $t_{Temp}\leftarrow$ Calculate the temporary start time of $o_{i+1}$ after $o_c$;
\State $t_2\leftarrow t_{Temp}-p_{i+1}$;
\If{$t_1>e_c \vee t_2> $ time slack of $o_{i+1}$}
\State \textbf{Continue};
\Else
\If{$t_1 \le d_c \wedge t_2 \le $ due time slack of $o_{i+1}$}
\State $Position_i.SetupIncrease \leftarrow s_{ic}+s_{c({i+1})}-s_{i({i+1})} $ ;
\State Add $Position_i$ into position list $PL1$;
\Else
\State $Position_i.Fitness \leftarrow$ Calculate the total fitness if $o_c$ is inserted;
\If{$Position_i.Fitness> f(S_C)$}
\State Add $Position_i$ into position list $PL2$;
\EndIf
\EndIf
\EndIf
\EndIf
\EndFor
\If{$PL1\neq \emptyset$}
\State $BestPosition\leftarrow$ Select the position increasing minimum setup time;
\State Insert $o_c$ into $S_C$ at $BestPosition$;
\State Update start times and time slacks;
\Else
\If{$PL2\neq \emptyset$}
\State $BestPosition\leftarrow$ Select the position increasing maximum fitness;
\State Insert $o_c$ into $S_C$ at $BestPosition$;
\State Update start times and time slacks;
\EndIf
\EndIf
\State \textbf{return} $S_C$;
\end{algorithmic}
\end{algorithm}

\textbf{Time slack and due time slack.} We set all the orders to start as early as possible. Therefore when inserting one order into the current solution at a position, it is possible to create more space for the candidate order by postponing some orders in the solution. In order to determine how much one order can be postponed, we adopt the \textit{time slack} idea from Verbeeck et al.~\cite{verbeeck2017time}. We further propose the \textit{due time slack} heuristic for this problem with tardiness penalty. The time slack is defined as the maximum amount of time an order can be postponed before the solution becomes infeasible. The time slack of each order depends on the latest start time of its succeeding order. Thus it is calculated from the last order to the first one in a back-propagation manner. The due time slack is the maximum amount of time an order can be postponed without adding penalty to any order. These heuristics facilitate determining the feasibility and the cost of one insertion only by comparing the time needed with the corresponding slack.

\textbf{Select the best position to insert.} For every candidate order, we calculate all possible insertion positions by comparing its time window with the current solution. For each possible solution, we do the following evaluation: Suppose we are evaluating the position between order $o_i$ and $o_{i+1}$ for the candidate order $o_c$, we first calculate the end time of $o_c$ if it is inserted after $o_i$ (denoted as $t_1$) and the time $o_{i+1}$ needed to be postponed (denoted as $t_2$). If $t_1$ is bigger than the deadline of $o_c$ or $t_2$ is bigger than the time slack of $o_{i+1}$, the position is given up; if $t_1$ is smaller than the due time of $o_c$ and $t_2$ is smaller than the due time slack (note that the due time slack is always smaller than the time slack), we calculate the increase of setup time if inserting $o_c$, which is $s_{ic}+s_{c({i+1})}-s_{i({i+1})}$ and add the position to candidate position list 1, $PL_1$; otherwise (i.e., if $t_1$ is larger than the due time of $o_c$ or $t_2$ is larger than the due time slack), we calculate the total fitness of the solution if we insert the order at this position and if it increases the fitness, we add the position to the candidate position list 2, $PL_2$. Finally, if $PL_1$ is not empty, the position with the smallest value of the increase of setup time in $PL_1$ is selected to insert the order; otherwise, the position with the highest total fitness in $PL_2$ is selected to insert the order. If both $PL_1$ and $PL_2$ are empty, the candidate order is given up.

We select the position according to the above strategy because when $PL_1$ is not empty, the candidate order can be inserted without receiving any penalty, which means the total fitness can be increased with the revenue of the candidate order. In this case we select the position increasing the minimum setup time. The rationale is that it is better to use the time more for processing orders instead of setting up. If $PL_1$ is empty, some orders will receive penalty. In this case we have to compute the fitness to find the best insertion position.

When an order is inserted, the start times of all its succeeding orders are updated until one whose start time does not change. The time slacks of all the orders before the candidate order and the orders whose start time is changed are also updated.

\section{Experiments}\label{sec_exp}

The goal of the experiments is to understand the strengths and weaknesses of the different algorithms for the OAS problem with sequence-dependent setup times and time windows. In particular, for the new algorithm, Sparrow, also to find out good parameter settings. For this purpose we start by running Sparrow for a range of parameter settings on a standard benchmark set. Then we compare Sparrow with state-of-the-art algorithms. Next we study how different properties of the problem correlate with its difficulty. Finally we generate new problem instances with more realistic and harder properties and compare the performances of different algorithms on the harder cases.

\subsection{Parameter settings for Sparrow}\label{sec_para}

In this section, we aim to understand how some important parameters of Sparrow influence its performance and answer the following questions: Is this hybridization better than the two standalone algorithms; for instances with different properties, is it better to have a larger population size or a larger number of iterations?

The standard benchmark set used is that by Cesaret et al.~\cite{cesaret2012tabu}, which is the most common benchmark set used by many articles \cite{cesaret2012tabu,nguyen2014enhancing,silva2018exact,lin2013increasing,chen2014diversity,nguyen2015dispatching,park2013evolving,park2014enhancing,nguyen2014sequential,nguyen2016learning,chaurasia2017hybrid,chaurasia2019artificial}. This benchmark set has $n$, $\tau$ and $r$ as parameters, where $n$ is the number of orders $n=25, 50, 100$, and $\tau$ and $R$ are the tardiness factor and the due time range factor, having the values of 0.1, 0.3, 0.5, 0.7, and 0.9. The orders are generated by random values from a discrete random distribution [1, 20] for processing time $t_i$ and revenue $r_i$, similarly the range [1, 10] is used for the sequence-dependent setup times $s_{ij}$. Release times $b_i$ of orders are from $[0, \tau\cdot
t_T]$, where $t_T=\sum_{i=1}^n t_i$. The due time for an order is calculated by $d_i=b_i+s_{\max}+\max\{v,t_i\}$, where $s_{\max}$ is the largest sequence-dependent setup time over any order, and $v$ is a random value generated from $[t_T(1-\tau-R/2),t_T(1-\tau+R/2) ]$. Finally the deadlines and weights are calculated by $e_i=d_i+Rt_i$ and $w_i=r_i/(e_i-d_i)$. For each set of parameters, 10 instances are generated.


Sparrow has a lot of parameters. In this section we focus on three parameters: the size of population $p$, the maximum number of consecutive iterations of no improvement and the maximum iteration, because these three parameters influence the importance of the two components in the hybridization. When the population size is small and the number of iterations is large, solutions are improved by more ALNS operations, hence the algorithm performance relies more on ALNS; when the population size is large and the number of iterations is small, the algorithm explores more solutions in the space, hence the algorithm performance relies more on BRKGA. We compare five sets of parameters, shown in Table~\ref{tab7}.  We change multiple parameters simultaneously and set them as in Table~\ref{tab7} because the Sparrow algorithm with these sets of parameters have relatively equal CPU time. We hope to see with equal CPU time, whether we should have larger population or more iterations for the Sparrow algorithm. In Table~\ref{tab7}, Set 1 refers to the case of standard ALNS and Set 5 refers to the case of standard BRKGA.

\begin{table}[htbp]
\scriptsize
  \centering
  \caption{Five sets of experiments with different parameters}
    \begin{tabular}{llllll}
    \toprule
    Parameter name& Set 1& Set 2& Set 3& Set 4 &Set 5\\
    \midrule
    Size of population $p$ & 1 & 10 & 20 & 50 & 1,000\\
    Maximum iteration of no improvement & 200$n$ & 20$n$ & 10$n$ & 4$n$ & $n$\\
    Maximum iteration & 1,000,000 & 100,000 & 50,000 & 20,000 & 2,000\\
    Whether ALNS is used &Yes&Yes&Yes&Yes&No\\
    \bottomrule
    \end{tabular}%
  \label{tab7}%
\end{table}%

Other parameters of Sparrow are set as follows: number of elite individuals $p_e=0.5p$; number of mutation individuals $p_m=0.1p$; probability for a gene coming from the elite parent $\rho_e=0.7$; the unit reward for a good pair $f_g=2$; probability of keeping the good pair $p_g=0.95$; number of orders to remove by ALNS is $0.4\cdot|S|$, where $|S|$ is the number of orders in the current solution; weight update parameter $\lambda=0.5$; coefficient of annealing $c_a=0.9975$; score increment according to the performance of operators: $\sigma_1=30,\sigma_2=20,\sigma_3=10$; parameter for ALNS improvement $p_f=0.9$. These parameters are set empirically through some exploratory experiments. Although these parameters are not optimal for all the instances, they provide relatively good solutions.

In this section, we compare the gaps of the results of Sparrow with the different parameter settings to the upper bounds provided by Silva et al.~\cite{silva2018exact}, because these upper bounds are tight and more suitable for analyzing the performance of an algorithm. We received the detailed upper bounds from the authors of \cite{silva2018exact}.

\begin{figure}[htbp]
\centering
\includegraphics[height=1.75in]{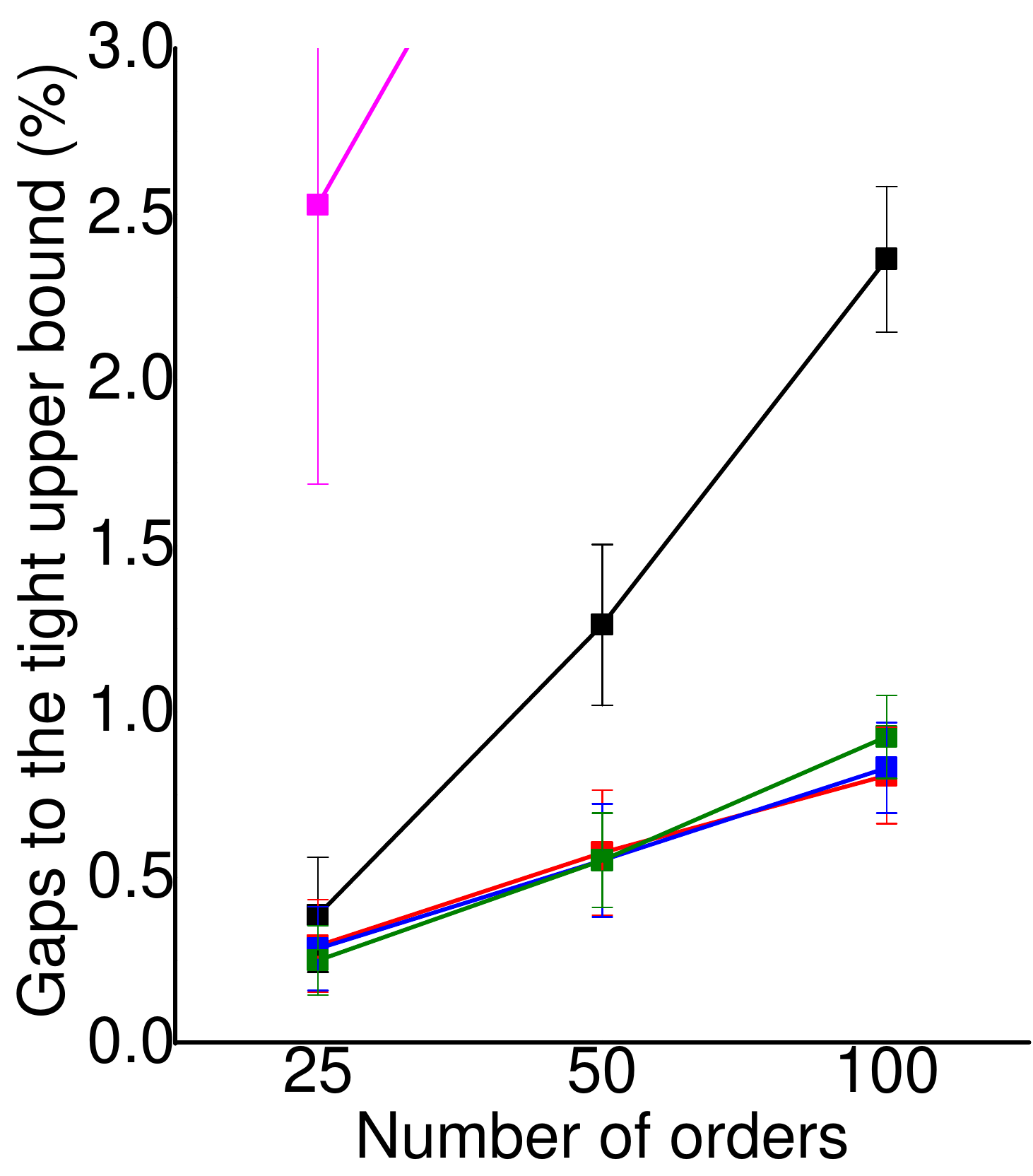}
\hfill
\includegraphics[height=1.75in]{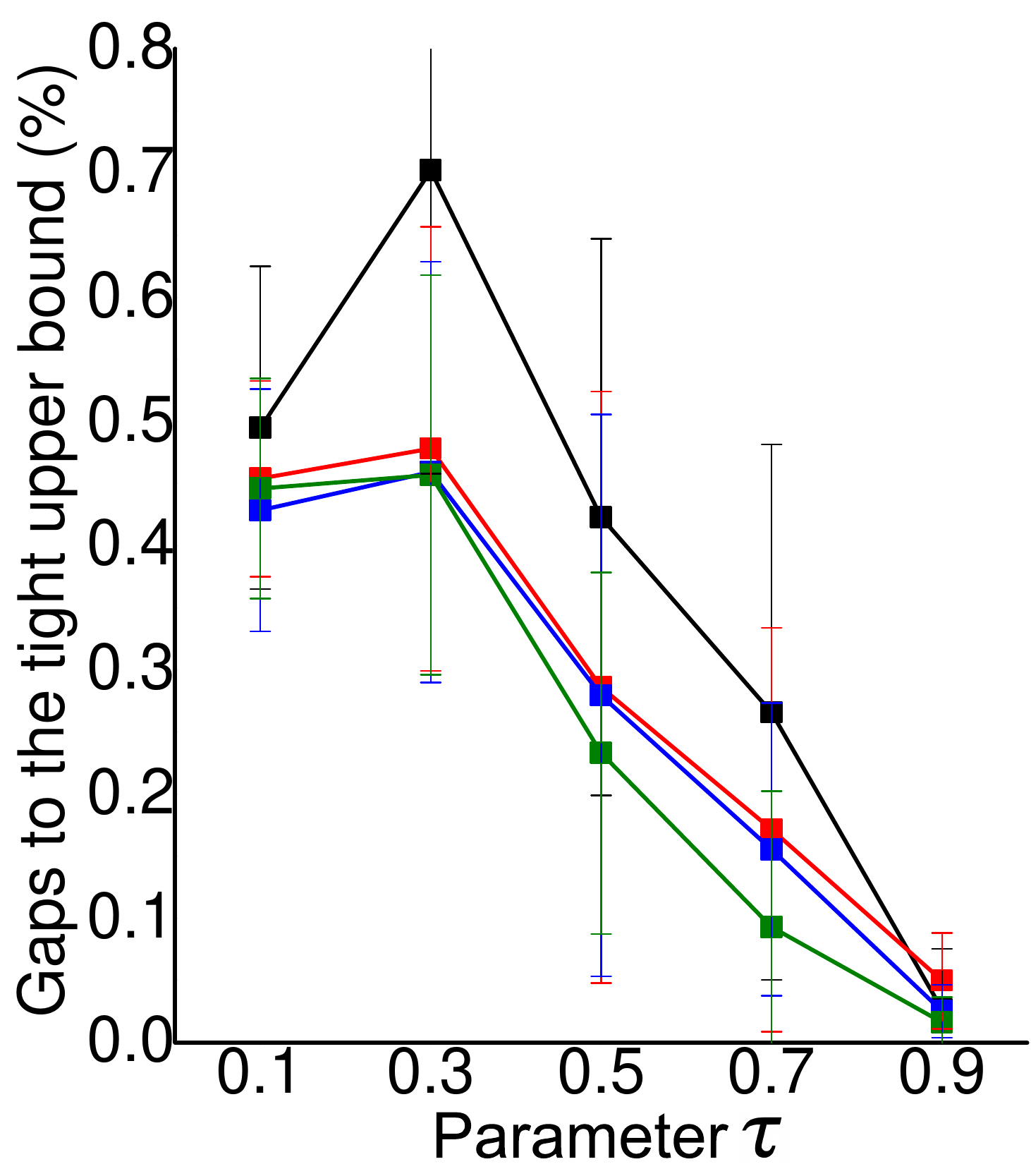}
\hfill
\includegraphics[height=1.75in]{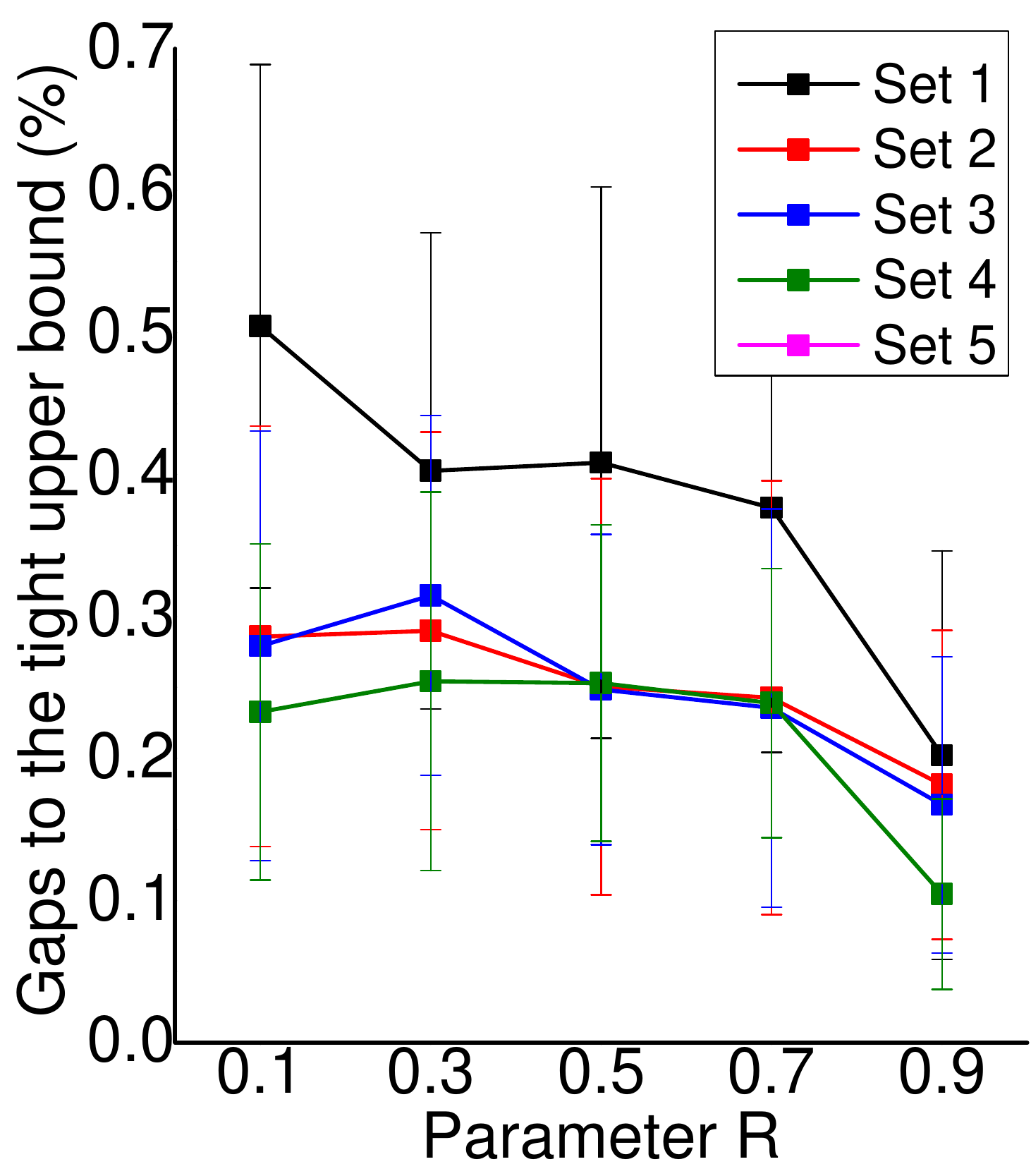}
\caption{The gap between the upper bound and Sparrow with different parameter settings. In Figure~\ref{figpara} left, each point is the average of 10 runs of all instances with the same number of orders in the standard benchmark set. In Figure~\ref{figpara} middle and right, we notice that the upper bounds from \cite{silva2018exact} are tighter for instances with 25 orders. Therefore, to better analyze the influence of $\tau$ and $r$, we calculate the average of 10 runs of instances with 25 orders and the same $\tau$ and $r$. Set 5 is quite literally ``off the charts''.}\label{figpara}
\end{figure}

The gaps of the Sparrow algorithm are calculated according to the different numbers of orders, $\tau$ and $R$ and shown in Figure~\ref{figpara}. We can only observe one point of results for Set 5 with the current scale, because results for Set 5 are significantly worse (e.g., 4.04\% for order size 50 and 6.00\% for order size 100). From all of the above results we conclude that the performance of the two standalone algorithms perform much worse than the hybrid algorithm, which proves the effectiveness of this hybridization. Comparing these three sub-figures, we can find that the Sparrow algorithm works better when the problem size is smaller, and when $\tau$ and $R$ are larger. We can also conclude that when the problem grows in size, and when $\tau$ and $R$ become smaller, the problem becomes harder for the Sparrow algorithm.

Then we compare Sets 2--4 within each sub-figure. Regarding the number of orders, Set 2 performs better when the problem size is larger, while Set 4 performs better when the problem size is smaller. The performance of Set 3 is between the performances of Set 2 and Set 4. It shows that the ALNS component is more suitable for larger instances, while the BRKGA component works better for smaller instances. Regarding the values of $\tau$ and $R$, the pattern is not very obvious. Set 4 tends to work best when $\tau$ and $R$ are larger.

As a summary, in this section we derive the following conclusions: (1) the hybrid Sparrow algorithm outperforms the two standalone algorithms; (2) the problem becomes harder for Sparrow when the problem grows in size and when $\tau$ and $R$ are smaller; (3) When the problem size is small, it is better to have a large population, while when the problem size is large, it is better to have a large iteration time; (4) Sparrow with a large population tends to work better when $\tau$ and $R$ are larger.

\subsection{Comparison with state-of-the-art algorithms}\label{sec_comparison}

In this section, we use the same benchmark set as mentioned in Section~\ref{sec_para}. We compare Sparrow to TS~\cite{cesaret2012tabu}, DRGA~\cite{nguyen2015dispatching}, GA~\cite{nguyen2016learning}, HH~\cite{nguyen2016learning}, LOS~\cite{nguyen2016learning}, ILS~\cite{silva2018exact}, ABC~\cite{lin2013increasing}, HSSGA~\cite{chaurasia2017hybrid}, and EA/G-LS~\cite{chaurasia2017hybrid}. Note that the performance of MIP solved by ILOG CPLEX has been tested by Cesaret et al.~\cite{cesaret2012tabu} and its performance is bad for large instances with more than 25 orders. Therefore, we do not compare Sparrow to CPLEX in this article.

The benchmark set has instances with the number of orders ranging from 25 to 100, and $\tau$ and $R$ ranging from 0.1 to 0.9. According to the conclusions derived in Section~\ref{sec_para}, we select a moderate set of parameters: population size: $p=20$; maximum number of consecutive iterations of no improvement: 10$n$; maximum iteration: 50,000. Other parameters are set as mentioned in Section~\ref{sec_para}.

Sparrow is run on Intel Core i5 3.20GHz CPU with 8GB memory, using a single core. For other methods, we present the results from the respective articles, which were obtained using machines with Intel Core i5, i7 and Xeon CPU, 3.00--3.40GHz, 4--16GB memory. Runtime results from such different machines are incomparable, even unmentioned details such as cache size can have a significant effect. However, we still include the runtime results for order size 100 in Table~\ref{tab_time}, to be able to draw some first conclusions about runtimes anyway.

On average, all the methods except ILS have comparable performance in terms of CPU time. ILS is slower than others. Regarding the solution quality, the above articles reported the gaps between their methods and the upper bounds provided by Cesaret et al.~\cite{cesaret2012tabu}. In order to compare with these methods directly, in this section we also use the upper bounds by Cesaret et al.~\cite{cesaret2012tabu} to calculate the gaps of Sparrow.

\begin{table}[htbp]
\tiny
  \centering
  \caption{The CPU time (s) for instances with 100 orders by different algorithms}
    \begin{tabular}{llllllllll}
    \toprule
    $\tau$ & R     & TS    & DRGA  & LOS   & ILS  & ABC   & HSSGA & EA/G-LS & Sparrow \\
    \midrule
    0.10  & 0.10  & 16.76  & 15    & 11    & 24.9  & \textbf{3.98 } & 17.91  & 13.72  & 10.55  \\
          & 0.30  & 17.26  & 15    & 11    & 25.3  & \textbf{8.22 } & 16.68  & 13.76  & 11.22  \\
          & 0.50  & 10.87  & 15    & 11    & 21.9  & \textbf{7.61 } & 17.07  & 12.82  & 7.88  \\
          & 0.70  & 6.21  & 15    & 11    & 13.9  & 6.25  & 15.08  & 11.06  & \textbf{1.38 } \\
          & 0.90  & 3.53  & 15    & 11    & 9.6   & 9.08  & 14.73  & 10.02  & \textbf{0.48 } \\
    0.30  & 0.10  & 22.37  & 15    & 11    & 38.4  & \textbf{4.89 } & 16.13  & 13.53  & 10.83  \\
          & 0.30  & 20.10  & 15    & 11    & 32.4  & \textbf{5.29 } & 19.70  & 14.33  & 10.75  \\
          & 0.50  & 16.77  & 15    & 11    & 33.7  & \textbf{4.88 } & 21.00  & 13.31  & 13.34  \\
          & 0.70  & 9.48  & 15    & 11    & 28.3  & \textbf{6.16 } & 16.60  & 14.04  & 9.16  \\
          & 0.90  & 7.58  & 15    & 11    & 25.1  & 8.59  & 18.33  & 12.10  & \textbf{6.75 } \\
    0.50  & 0.10  & 25.96  & 15    & 12    & 51.9  & \textbf{5.27 } & 19.30  & 12.24  & 11.05  \\
          & 0.30  & 28.87  & 15    & 12    & 57.9  & \textbf{6.94 } & 20.96  & 12.87  & 12.21  \\
          & 0.50  & 20.56  & 15    & 12    & 52.5  & \textbf{5.86 } & 21.52  & 13.75  & 15.05  \\
          & 0.70  & 15.57  & 15    & 12    & 42.7  & \textbf{6.57 } & 17.27  & 11.79  & 15.49  \\
          & 0.90  & 12.15  & 15    & 12    & 46.0  & \textbf{6.76 } & 18.66  & 13.72  & 16.89  \\
    0.70  & 0.10  & 33.60  & 15    & 13    & 63.9  & \textbf{6.04 } & 23.40  & 11.75  & 16.46  \\
          & 0.30  & 26.62  & 15    & 12    & 73.9  & \textbf{6.25 } & 24.90  & 12.79  & 17.98  \\
          & 0.50  & 22.30  & 15    & 12    & 72.6  & \textbf{6.77 } & 19.89  & 12.79  & 18.86  \\
          & 0.70  & 26.36  & 16    & 12    & 77.5  & \textbf{6.78 } & 24.24  & 14.33  & 20.53  \\
          & 0.90  & 17.84  & 16    & 12    & 78.4  & \textbf{7.42 } & 21.90  & 15.45  & 18.78  \\
    0.90  & 0.10  & 29.46  & 16    & 12    & 80.6  & 13.63  & 19.59  & \textbf{10.65 } & 13.92  \\
          & 0.30  & 26.32  & 16    & 12    & 79.1  & \textbf{9.82 } & 19.64  & 11.75  & 13.75  \\
          & 0.50  & 21.51  & 16    & 12    & 81.1  & \textbf{6.46 } & 19.06  & 11.62  & 15.48  \\
          & 0.70  & 22.70  & 16    & 12    & 91.5  & \textbf{7.52 } & 17.61  & 14.30  & 16.69  \\
          & 0.90  & 17.48  & 16    & 12    & 92.0  & \textbf{8.61 } & 17.92  & 11.04  & 17.38  \\
    Avg.  &       & 19.13  & 15    & 11    & 51.8  & \textbf{7.03 } & 19.16  & 12.78  & 12.91  \\
    \bottomrule
    \end{tabular}%
  \label{tab_time}%
\end{table}%

\begin{sidewaystable}[htbp]
\tiny
  \centering
  \caption{Gaps (\%) of various algorithms on instances with 25 orders (multiple runs for each instance) \protect\tablefootnote{For the gaps of instances with 25 orders reported by Silva et al.~\cite{silva2018exact}, we identify some mistakes because the gaps in their table are smaller than the gaps between their tight upper bounds and the upper bounds by Cesaret et al.~\cite{cesaret2012tabu}. Therefore we decide not to include ILS in this comparison.}
}
    \begin{tabular}{lllllllllllllllllllllllll}
    \toprule
    \multicolumn{2}{l}{n=25} & \multicolumn{3}{l}{TS} &       & \multicolumn{3}{l}{DRGA} &       & \multicolumn{3}{l}{GA} &       & \multicolumn{3}{l}{HH} &       & \multicolumn{3}{l}{LOS} &       & \multicolumn{3}{l}{Sparrow} \\
\cmidrule{1-5}\cmidrule{7-9}\cmidrule{11-13}\cmidrule{15-17}\cmidrule{19-21}\cmidrule{23-25}    $\tau$ & R     & Min   & Avg   & Max   &       & Min   & Avg   & Max   &       & Min   & Avg   & Max   &       & Min   & Avg   & Max   &       & Min   & Avg   & Max   &       & Min   & Avg   & Max \\
    \midrule
    0.10  & 0.10  & 1     & 4     & 6     &       & 2     & 3     & 4     &       & 1     & 2     & 3     &       & 3     & 5     & 8     &       & 1     & 2     & 3     &       & \textbf{0.72} & \textbf{1.82} & \textbf{2.69} \\
          & 0.30  & 2     & 3     & 6     &       & 1     & 2     & 6     &       & 1     & 2     & 5     &       & 2     & 4     & 9     &       & \textbf{0} & 2     & 5     &       & \textbf{0.68} & \textbf{1.53} & \textbf{3.19} \\
          & 0.50  & 1     & 2     & 4     &       & 1     & 1     & 3     &       & \textbf{0} & 1     & 2     &       & 1     & 4     & 7     &       & \textbf{0} & 1     & 3     &       & \textbf{0.00} & \textbf{0.71} & \textbf{1.60} \\
          & 0.70  & \textbf{0} & 1     & 4     &       & \textbf{0} & 1     & 3     &       & \textbf{0} & \textbf{0} & 2     &       & \textbf{0} & 2     & 5     &       & \textbf{0} & \textbf{0} & 2     &       & \textbf{0.00} & \textbf{0.33} & \textbf{1.27} \\
          & 0.90  & \textbf{0} & 1     & \textbf{2} &       & \textbf{0} & 1     & \textbf{2} &       & \textbf{0} & \textbf{0} & \textbf{2} &       & \textbf{0} & 2     & 5     &       & \textbf{0} & \textbf{0} & \textbf{2} &       & \textbf{0.00} & \textbf{0.34} & \textbf{2.08} \\
    0.30  & 0.10  & 2     & 4     & 5     &       & 2     & 3     & 5     &       & 1     & 3     & \textbf{4} &       & 3     & 7     & 10    &       & 1     & 3     & \textbf{4} &       & \textbf{0.89} & \textbf{2.31} & \textbf{4.19} \\
          & 0.30  & 3     & 5     & 7     &       & 2     & 3     & 5     &       & \textbf{1} & 3     & \textbf{4} &       & 4     & 7     & 10    &       & 2     & 3     & 5     &       & \textbf{1.11} & \textbf{2.94} & 5.06 \\
          & 0.50  & 2     & 3     & 6     &       & 2     & 2     & \textbf{3} &       & \textbf{0} & 2     & \textbf{3} &       & 3     & 4     & 9     &       & \textbf{0} & 2     & \textbf{3} &       & \textbf{0.86} & \textbf{1.65} & \textbf{3.49} \\
          & 0.70  & 1     & 2     & 6     &       & 1     & 2     & 5     &       & \textbf{0} & \textbf{1} & 5     &       & 2     & 4     & 10    &       & \textbf{0} & \textbf{1} & 5     &       & \textbf{0.00} & \textbf{1.69} & \textbf{4.66} \\
          & 0.90  & \textbf{0} & 2     & 4     &       & \textbf{0} & \textbf{1} & 3     &       & \textbf{0} & \textbf{1} & \textbf{2} &       & \textbf{0} & 4     & 7     &       & \textbf{0} & \textbf{1} & \textbf{2} &       & \textbf{0.00} & \textbf{1.06} & \textbf{2.79} \\
    0.50  & 0.10  & 3     & 6     & 7     &       & 2     & 4     & 7     &       & \textbf{1} & 4     & 6     &       & 4     & 8     & 11    &       & \textbf{1} & 4     & \textbf{5} &       & \textbf{1.68} & \textbf{3.87} & 6.14 \\
          & 0.30  & 3     & 5     & 9     &       & 2     & 5     & 8     &       & 2     & 4     & \textbf{7} &       & 4     & 8     & 13    &       & \textbf{1} & 4     & \textbf{7} &       & \textbf{1.79} & \textbf{3.99} & \textbf{7.16} \\
          & 0.50  & 2     & 5     & 8     &       & 1     & 5     & \textbf{6} &       & 1     & \textbf{4} & \textbf{6} &       & 5     & 8     & 11    &       & 1     & \textbf{4} & \textbf{6} &       & \textbf{0.97} & \textbf{4.32} & \textbf{6.31} \\
          & 0.70  & 2     & 6     & 11    &       & 2     & \textbf{4} & \textbf{7} &       & \textbf{0} & \textbf{4} & \textbf{7} &       & 2     & 7     & 11    &       & \textbf{0} & \textbf{4} & \textbf{7} &       & \textbf{0.63} & \textbf{4.26} & \textbf{7.29} \\
          & 0.90  & 1     & 4     & 7     &       & 1     & \textbf{3} & 7     &       & 1     & \textbf{3} & 7     &       & 3     & 6     & 9     &       & 1     & \textbf{3} & \textbf{6} &       & \textbf{0.85} & \textbf{3.10} & 7.00 \\
    0.70  & 0.10  & 3     & 9     & 18    &       & 1     & 8     & 16    &       & \textbf{0} & \textbf{7} & \textbf{14} &       & 6     & 11    & 19    &       & \textbf{0} & \textbf{7} & 15    &       & \textbf{0.92} & \textbf{7.46} & 15.81 \\
          & 0.30  & 7     & 10    & 14    &       & \textbf{5} & 9     & 13    &       & \textbf{5} & \textbf{8} & \textbf{12} &       & 10    & 12    & 16    &       & \textbf{5} & \textbf{8} & \textbf{12} &       & \textbf{5.34} & \textbf{8.43} & \textbf{12.39} \\
          & 0.50  & 7     & 12    & 15    &       & 6     & \textbf{10} & \textbf{14} &       & \textbf{5} & \textbf{10} & \textbf{14} &       & 10    & 14    & 18    &       & \textbf{5} & \textbf{10} & \textbf{14} &       & \textbf{5.41} & \textbf{10.11} & \textbf{14.02} \\
          & 0.70  & 2     & 8     & 14    &       & 2     & 7     & \textbf{12} &       & \textbf{1} & \textbf{6} & \textbf{12} &       & \textbf{1} & 9     & 14    &       & \textbf{1} & \textbf{6} & \textbf{12} &       & \textbf{1.69} & \textbf{6.80} & \textbf{12.08} \\
          & 0.90  & 3     & 10    & 15    &       & 1     & \textbf{8} & 14    &       & \textbf{0} & \textbf{8} & \textbf{13} &       & 2     & 11    & 15    &       & \textbf{0} & \textbf{8} & \textbf{13} &       & \textbf{0.65} & \textbf{8.21} & \textbf{13.63} \\
    0.90  & 0.10  & \textbf{0} & 1     & 6     &       & \textbf{0} & 1     & 5     &       & \textbf{0} & 1     & 5     &       & \textbf{0} & 1     & 6     &       & \textbf{0} & 1     & 5     &       & \textbf{0.00} & \textbf{0.49} & \textbf{4.92} \\
          & 0.30  & \textbf{0} & 1     & 1     &       & \textbf{0} & 1     & 1     &       & \textbf{0} & \textbf{0} & \textbf{0} &       & \textbf{0} & 2     & 7     &       & \textbf{0} & \textbf{0} & \textbf{0} &       & \textbf{0.00} & \textbf{0.00} & \textbf{0.01} \\
          & 0.50  & \textbf{0} & 4     & \textbf{12} &       & \textbf{0} & \textbf{2} & \textbf{12} &       & \textbf{0} & \textbf{2} & \textbf{12} &       & 1     & 5     & 18    &       & \textbf{0} & \textbf{2} & \textbf{12} &       & \textbf{0.00} & \textbf{2.56} & \textbf{12.30} \\
          & 0.70  & 1     & 8     & 25    &       & 1     & 7     & 21    &       & \textbf{0} & \textbf{6} & 21    &       & \textbf{0} & 9     & 22    &       & \textbf{0} & \textbf{6} & 21    &       & \textbf{0.00} & \textbf{6.64} & \textbf{20.99} \\
          & 0.90  & 1     & 7     & 22    &       & 1     & \textbf{6} & 20    &       & \textbf{0} & \textbf{6} & \textbf{19} &       & \textbf{0} & 7     & 21    &       & \textbf{0} & \textbf{6} & \textbf{19} &       & \textbf{0.00} & \textbf{6.02} & \textbf{19.16} \\
    Avg.  &       & 2     & 5     & 9     &       & 1     & 3     & 8     &       & \textbf{0} & \textbf{3} & \textbf{7} &       & 2     & 6     & 11    &       & \textbf{0} & \textbf{3} & \textbf{7} &       & \textbf{0.97} & \textbf{3.63} & \textbf{7.61} \\
    \bottomrule
    \end{tabular}%
  \label{tab1}%
\end{sidewaystable}%

\begin{sidewaystable}[htbp]
\tiny
  \centering
  \caption{Gaps (\%) of various algorithms on instances with 50 orders (multiple runs for each instance)\protect\tablefootnote{For the gaps of instances with 50 orders reported by Silva et al.~\cite{silva2018exact}, we identify some mistakes because the gaps in their table are smaller than the gaps between their tight upper bounds and the upper bounds by Cesaret et al.~\cite{cesaret2012tabu}. Therefore we decide not to include ILS in this comparison.}
}
    \begin{tabular}{lllllllllllllllllllllllll}
    \toprule
    \multicolumn{2}{l}{n=50} & \multicolumn{3}{l}{TS} &       & \multicolumn{3}{l}{DRGA} &       & \multicolumn{3}{l}{GA} &       & \multicolumn{3}{l}{HH} &       & \multicolumn{3}{l}{LOS} &       & \multicolumn{3}{l}{Sparrow} \\
\cmidrule{1-5}\cmidrule{7-9}\cmidrule{11-13}\cmidrule{15-17}\cmidrule{19-21}\cmidrule{23-25}    $\tau$ & R     & Min   & Avg   & Max   &       & Min   & Avg   & Max   &       & Min   & Avg   & Max   &       & Min   & Avg   & Max   &       & Min   & Avg   & Max   &       & Min   & Avg   & Max \\
    \midrule
    0.10  & 0.10  & 1     & 2     & 3     &       & 1     & 2     & 3     &       & 1     & 2     & 3     &       & 3     & 4     & 5     &       & 1     & 2     & 3     &       & \textbf{0.51} & \textbf{0.79} & \textbf{1.27} \\
          & 0.30  & 1     & 2     & 4     &       & 1     & 2     & 3     &       & 1     & 2     & 4     &       & 2     & 4     & 5     &       & 1     & 2     & 3     &       & \textbf{0.60} & \textbf{1.00} & \textbf{1.53} \\
          & 0.50  & 1     & 2     & 2     &       & 1     & 1     & 2     &       & \textbf{0} & 1     & 2     &       & 1     & 2     & 4     &       & \textbf{0} & 1     & 2     &       & \textbf{0.00} & \textbf{0.42} & \textbf{0.96} \\
          & 0.70  & \textbf{0} & 2     & \textbf{16} &       & \textbf{0} & 2     & \textbf{16} &       & \textbf{0} & 2     & \textbf{16} &       & \textbf{0} & 3     & 17    &       & \textbf{0} & 2     & \textbf{16} &       & \textbf{0.00} & \textbf{1.66} & \textbf{16.38} \\
          & 0.90  & \textbf{0} & 1     & 2     &       & \textbf{0} & \textbf{0} & \textbf{0} &       & \textbf{0} & \textbf{0} & \textbf{0} &       & \textbf{0} & \textbf{0} & 1     &       & \textbf{0} & \textbf{0} & \textbf{0} &       & \textbf{0.00} & \textbf{0.00} & \textbf{0.00} \\
    0.30  & 0.10  & 2     & 3     & 3     &       & 2     & 3     & 4     &       & 1     & 2     & 4     &       & 3     & 6     & 8     &       & 1     & 2     & 3     &       & \textbf{0.63} & \textbf{1.59} & \textbf{2.86} \\
          & 0.30  & 3     & 4     & 5     &       & 2     & 3     & 4     &       & 2     & 3     & 4     &       & 5     & 6     & 9     &       & 2     & 3     & 4     &       & \textbf{1.40} & \textbf{1.81} & \textbf{2.69} \\
          & 0.50  & 1     & 3     & 5     &       & 1     & 2     & 4     &       & 1     & 2     & 4     &       & 1     & 4     & 7     &       & 1     & 2     & \textbf{3} &       & \textbf{0.18} & \textbf{1.44} & \textbf{3.64} \\
          & 0.70  & \textbf{0} & 1     & 3     &       & \textbf{0} & 1     & 2     &       & \textbf{0} & \textbf{0} & \textbf{1} &       & 1     & 3     & 5     &       & \textbf{0} & 1     & 2     &       & \textbf{0.00} & \textbf{0.39} & \textbf{1.16} \\
          & 0.90  & 1     & 1     & 3     &       & \textbf{0} & 1     & 2     &       & \textbf{0} & \textbf{0} & \textbf{1} &       & \textbf{0} & 2     & 4     &       & \textbf{0} & \textbf{0} & \textbf{1} &       & \textbf{0.00} & \textbf{0.27} & \textbf{1.32} \\
    0.50  & 0.10  & 3     & 4     & 5     &       & 3     & 4     & 5     &       & \textbf{1} & \textbf{2} & 3     &       & 4     & 6     & 8     &       & \textbf{1} & 3     & 4     &       & \textbf{1.11} & \textbf{2.01} & \textbf{2.64} \\
          & 0.30  & 3     & 6     & 8     &       & 3     & 5     & 7     &       & 2     & \textbf{3} & 6     &       & 6     & 8     & 11    &       & 2     & 4     & 6     &       & \textbf{1.97} & \textbf{3.30} & \textbf{5.44} \\
          & 0.50  & 2     & 4     & 8     &       & 2     & 4     & 8     &       & \textbf{1} & \textbf{3} & \textbf{6} &       & 3     & 7     & 12    &       & \textbf{1} & \textbf{3} & 7     &       & \textbf{1.03} & \textbf{3.00} & \textbf{6.49} \\
          & 0.70  & 2     & 3     & 5     &       & 1     & 3     & 6     &       & \textbf{0} & 2     & \textbf{4} &       & 2     & 5     & 9     &       & \textbf{0} & 2     & \textbf{4} &       & \textbf{0.37} & \textbf{1.97} & \textbf{4.14} \\
          & 0.90  & 0     & 3     & 6     &       & 0     & 2     & 5     &       & 0     & \textbf{1} & \textbf{4} &       & 0     & 5     & 7     &       & 0     & \textbf{1} & \textbf{4} &       & \textbf{-4.09} & \textbf{1.53} & \textbf{4.04} \\
    0.70  & 0.10  & 4     & 7     & 9     &       & 4     & 5     & 7     &       & \textbf{2} & 4     & 5     &       & 7     & 9     & 11    &       & \textbf{2} & 4     & 5     &       & \textbf{2.24} & \textbf{3.88} & \textbf{4.74} \\
          & 0.30  & 4     & 7     & 11    &       & 4     & 7     & 10    &       & 3     & 5     & \textbf{8} &       & 8     & 10    & 14    &       & 3     & 5     & 9     &       & \textbf{2.75} & \textbf{4.82} & \textbf{8.40} \\
          & 0.50  & 7     & 9     & 13    &       & 6     & 8     & 12    &       & \textbf{4} & \textbf{6} & 11    &       & 8     & 11    & 15    &       & 5     & \textbf{6} & 11    &       & \textbf{4.74} & \textbf{6.46} & \textbf{10.91} \\
          & 0.70  & 2     & 9     & 18    &       & 3     & 8     & 15    &       & 2     & \textbf{7} & 15    &       & 6     & 11    & 20    &       & 2     & \textbf{7} & \textbf{14} &       & \textbf{1.73} & \textbf{7.13} & 15.50 \\
          & 0.90  & 4     & 10    & 18    &       & 4     & 9     & 14    &       & \textbf{3} & \textbf{7} & \textbf{13} &       & 5     & 10    & 16    &       & \textbf{3} & \textbf{7} & \textbf{13} &       & \textbf{3.26} & \textbf{7.63} & \textbf{13.02} \\
    0.90  & 0.10  & 8     & 13    & 18    &       & 8     & 12    & 19    &       & \textbf{6} & 11    & \textbf{16} &       & 10    & 14    & 21    &       & 7     & 11    & \textbf{16} &       & \textbf{6.55} & \textbf{10.98} & \textbf{16.77} \\
          & 0.30  & 12    & 16    & 21    &       & \textbf{9} & 14    & 18    &       & \textbf{9} & \textbf{13} & \textbf{17} &       & 13    & 17    & 21    &       & \textbf{9} & \textbf{13} & \textbf{17} &       & \textbf{9.28} & \textbf{13.43} & \textbf{17.80} \\
          & 0.50  & 7     & 16    & 20    &       & 4     & 13    & 19    &       & \textbf{3} & \textbf{12} & 17    &       & 5     & 16    & 20    &       & \textbf{3} & \textbf{12} & \textbf{16} &       & \textbf{3.05} & \textbf{12.31} & \textbf{16.78} \\
          & 0.70  & 8     & 14    & 21    &       & 7     & 13    & 18    &       & \textbf{5} & \textbf{11} & \textbf{17} &       & 7     & 15    & 22    &       & 6     & \textbf{11} & \textbf{17} &       & \textbf{5.62} & \textbf{11.50} & 18.42 \\
          & 0.90  & 11    & 14    & 19    &       & 10    & 13    & 17    &       & \textbf{9} & \textbf{12} & \textbf{16} &       & 12    & 16    & 20    &       & 10    & \textbf{12} & \textbf{16} &       & \textbf{9.74} & \textbf{12.49} & \textbf{16.30} \\
    Avg.  &       & 3     & 6     & 10    &       & 3     & 5     & 9     &       & \textbf{2} & 5     & 8     &       & 4     & 8     & 12    &       & \textbf{2} & 5     & 8     &       & \textbf{2.11} & \textbf{4.47} & \textbf{7.73} \\
    \bottomrule
    \end{tabular}%
  \label{tab2}%
  \end{sidewaystable}%

\begin{sidewaystable}[htbp]
\tiny
  \centering
  \caption{Gaps (\%) of various algorithms on instances with 100 orders (multiple runs for each instance)}
    \begin{tabular}{lllllllllllllllllllllllllllll}
    \toprule
    \multicolumn{2}{l}{n=100} & \multicolumn{3}{l}{TS} &       & \multicolumn{3}{l}{DRGA} &       & \multicolumn{3}{l}{GA} &       & \multicolumn{3}{l}{HH} &       & \multicolumn{3}{l}{LOS} &       & \multicolumn{3}{l}{ILS-Avg} &       & \multicolumn{3}{l}{Sparrow} \\
\cmidrule{1-5}\cmidrule{7-9}\cmidrule{11-13}\cmidrule{15-17}\cmidrule{19-21}\cmidrule{23-25}\cmidrule{27-29}    $\tau$ & R     & Min   & Avg   & Max   &       & Min   & Avg   & Max   &       & Min   & Avg   & Max   &       & Min   & Avg   & Max   &       & Min   & Avg   & Max   &       & Min   & Avg   & Max   &       & Min   & Avg   & Max \\
    \midrule
    0.10  & 0.10  & 1     & 2     & 3     &       & 1     & 1     & 2     &       & 1     & 2     & 3     &       & 2     & 3     & 5     &       & 2     & 2     & 3     &       & 0.74  & 0.95  & 1.35  &       & \textbf{0.39} & \textbf{0.57} & \textbf{0.80} \\
          & 0.30  & 2     & 2     & 3     &       & 1     & 1     & 2     &       & 2     & 2     & 3     &       & 1     & 3     & 4     &       & 1     & 2     & 3     &       & 0.44  & 0.74  & 1.09  &       & \textbf{0.34} & \textbf{0.61} & \textbf{0.88} \\
          & 0.50  & 1     & 1     & 3     &       & 1     & 1     & 1     &       & 1     & 1     & 2     &       & 1     & 1     & 2     &       & 1     & 1     & 2     &       & 0.20  & 0.37  & 0.58  &       & \textbf{0.00} & \textbf{0.13} & \textbf{0.28} \\
          & 0.70  & \textbf{0} & 1     & 1     &       & \textbf{0} & 1     & 1     &       & \textbf{0} & \textbf{0} & \textbf{0} &       & \textbf{0} & \textbf{0} & 2     &       & \textbf{0} & \textbf{0} & 1     &       & \textbf{0.00} & 0.04  & 0.12  &       & \textbf{0.00} & \textbf{0.00} & \textbf{0.00} \\
          & 0.90  & \textbf{0} & 1     & 1     &       & \textbf{0} & \textbf{0} & \textbf{0} &       & \textbf{0} & \textbf{0} & \textbf{0} &       & \textbf{0} & \textbf{0} & \textbf{0} &       & \textbf{0} & \textbf{0} & \textbf{0} &       & \textbf{0.00} & 0.01  & 0.10  &       & \textbf{0.00} & \textbf{0.00} & \textbf{0.00} \\
    0.30  & 0.10  & 1     & 3     & 4     &       & 2     & 3     & 4     &       & 2     & 3     & 4     &       & 4     & 6     & 7     &       & 2     & 2     & 3     &       & 1.00  & 1.40  & 1.82  &       & \textbf{0.53} & \textbf{0.99} & \textbf{1.44} \\
          & 0.30  & 2     & 3     & 5     &       & 1     & 3     & 4     &       & 1     & 2     & 4     &       & 2     & 5     & 7     &       & 1     & 3     & 4     &       & 0.73  & 1.38  & 2.67  &       & \textbf{0.57} & \textbf{1.24} & \textbf{2.49} \\
          & 0.50  & 1     & 2     & 4     &       & 2     & 2     & 3     &       & 1     & 2     & 3     &       & 3     & 4     & 6     &       & 1     & 2     & 3     &       & 0.73  & 1.17  & \textbf{1.65} &       & \textbf{0.65} & \textbf{1.07} & 1.75 \\
          & 0.70  & 1     & 2     & 3     &       & 1     & 1     & 1     &       & \textbf{0} & 1     & 2     &       & 1     & 2     & 4     &       & \textbf{0} & 1     & 2     &       & 0.27  & 0.44  & \textbf{0.74} &       & \textbf{0.00} & \textbf{0.30} & 0.79 \\
          & 0.90  & 1     & 1     & 2     &       & \textbf{0} & 1     & 1     &       & \textbf{0} & \textbf{0} & 1     &       & \textbf{0} & 1     & 3     &       & \textbf{0} & \textbf{0} & 1     &       & \textbf{0.00} & 0.25  & 0.73  &       & \textbf{0.00} & \textbf{0.11} & \textbf{0.42} \\
    0.50  & 0.10  & 2     & 4     & 5     &       & 3     & 5     & 7     &       & 3     & 4     & 6     &       & 7     & 8     & 11    &       & 2     & 4     & 5     &       & 1.46  & 2.26  & 3.11  &       & \textbf{1.07} & \textbf{1.79} & \textbf{2.71} \\
          & 0.30  & 3     & 4     & 6     &       & 4     & 4     & 6     &       & 3     & 4     & 5     &       & 6     & 7     & 9     &       & 2     & 3     & 5     &       & 1.72  & 2.32  & 3.08  &       & \textbf{1.49} & \textbf{2.16} & \textbf{2.63} \\
          & 0.50  & 3     & 4     & 5     &       & 3     & 4     & 6     &       & 2     & 4     & 5     &       & 5     & 7     & 10    &       & 2     & 3     & 4     &       & 1.58  & 2.40  & 3.36  &       & \textbf{1.31} & \textbf{2.33} & \textbf{3.35} \\
          & 0.70  & 2     & 3     & 4     &       & 1     & 3     & 4     &       & \textbf{0} & 2     & 3     &       & 2     & 5     & 7     &       & \textbf{0} & 2     & 3     &       & 0.49  & 1.61  & 2.80  &       & \textbf{0.35} & \textbf{1.49} & \textbf{2.38} \\
          & 0.90  & 1     & 2     & 5     &       & 1     & 2     & 4     &       & \textbf{0} & 1     & 3     &       & 2     & 3     & 6     &       & \textbf{0} & 1     & 4     &       & 0.39  & 1.16  & 2.76  &       & \textbf{0.19} & \textbf{0.92} & \textbf{2.40} \\
    0.70  & 0.10  & 3     & 5     & 6     &       & 4     & 6     & 8     &       & 4     & 5     & 6     &       & 7     & 9     & 11    &       & 3     & 4     & 7     &       & 2.28  & 3.13  & 3.77  &       & \textbf{1.69} & \textbf{2.47} & \textbf{3.21} \\
          & 0.30  & 4     & 7     & 11    &       & 4     & 6     & 9     &       & 3     & 5     & 8     &       & 6     & 9     & 13    &       & 2     & 5     & 7     &       & 1.89  & 3.86  & 5.82  &       & \textbf{1.35} & \textbf{3.66} & \textbf{5.80} \\
          & 0.50  & 4     & 6     & 13    &       & 4     & 7     & 15    &       & 3     & 6     & 12    &       & 6     & 10    & 18    &       & \textbf{2} & 5     & 12    &       & 2.63  & \textbf{4.25} & \textbf{8.69}\tablefootnote{For this value reported by Silva et al.~\cite{silva2018exact}, we find it smaller than the gaps between their tight upper bounds and the upper bounds by Cesaret et al.~\cite{cesaret2012tabu}.}
 &   & \textbf{2.23} & 4.34 & 10.65 \\
          & 0.70  & 3     & 7     & 13    &       & 5     & 8     & 12    &       & 4     & 6     & 9     &       & 6     & 9     & 13    &       & 2     & 5     & 8     &       & 2.66  & 6.17  & 9.32  &       & \textbf{1.92} & \textbf{4.81} & \textbf{7.33} \\
          & 0.90  & 5     & 8     & 13    &       & 4     & 7     & 10    &       & 4     & 6     & 9     &       & 5     & 9     & 12    &       & 3     & \textbf{5} & 8     &       & 3.92  & 6.60  & 8.27  &       & \textbf{2.97} & \textbf{5.07} & \textbf{6.55} \\
    0.90  & 0.10  & 7     & 9     & 12    &       & 6     & 9     & 12    &       & 5     & 7     & 9     &       & 8     & 11    & 14    &       & 4     & 6     & 9     &       & 4.40  & 7.02  & 9.07  &       & \textbf{3.44} & \textbf{5.46} & \textbf{8.47} \\
          & 0.30  & 7     & 14    & 17    &       & 8     & 12    & 16    &       & 6     & 10    & 13    &       & 11    & 14    & 18    &       & 6     & 9     & 12    &       & 8.91  & 11.83 & 13.59 &       & \textbf{4.90} & \textbf{8.65} & \textbf{11.20} \\
          & 0.50  & 11    & 16    & 18    &       & 11    & 14    & 19    &       & 8     & 12    & 17    &       & 12    & 16    & 20    &       & 8     & 11    & 16    &       & 10.12 & 14.06 & 18.40 &       & \textbf{7.90} & \textbf{10.76} & \textbf{15.67} \\
          & 0.70  & 11    & 15    & 20    &       & 10    & 14    & 16    &       & 9     & 12    & 15    &       & 11    & 16    & 19    &       & 8     & 11    & 14    &       & 9.67  & 12.75 & 15.95 &       & \textbf{7.38} & \textbf{10.73} & \textbf{13.02} \\
          & 0.90  & 11    & 16    & 22    &       & 8     & 13    & 19    &       & 6     & 12    & 17    &       & 8     & 15    & 22    &       & 4     & 11    & 18    &       & 7.21  & 13.23 & 18.43 &       & \textbf{3.32} & \textbf{10.67} & \textbf{16.53} \\
    Avg.  &       & 3     & 6     & 8     &       & 3     & 5     & 7     &       & 3     & 4     & 6     &       & 5     & 7     & 10    &       & 2     & 4     & 6     &       & 2.54  & 3.98  & 5.49  &       & \textbf{1.76} & \textbf{3.21} & \textbf{4.83} \\
    \bottomrule
    \end{tabular}%
  \label{tab3}%
\end{sidewaystable}%

\begin{sidewaystable}[htbp]
\tiny
  \centering
  \caption{Gaps (\%) of various algorithms on instances with 25 orders (single run for each instance)}
    \begin{tabular}{lllllllllllllllllllllllll}
    \toprule
    \multicolumn{2}{l}{n=25} & \multicolumn{3}{l}{ABC} &       & \multicolumn{3}{l}{HSSGA} &       & \multicolumn{3}{l}{EA/G-LS} &       & \multicolumn{3}{l}{Sparrow-Average} &       & \multicolumn{3}{l}{Sparrow-Best} &       & \multicolumn{3}{c}{Sparrow-Worst} \\
\cmidrule{1-5}\cmidrule{7-9}\cmidrule{11-13}\cmidrule{15-17}\cmidrule{19-21}\cmidrule{23-25}    $\tau$ & R     & Min   & Avg   & Max   &       & Min   & Avg   & Max   &       & Min   & Avg   & Max   &       & Min   & Avg   & Max   &       & Min   & Avg   & Max   &       & Min   & Avg   & Max \\
    \midrule
    0.10  & 0.10  & 1.45  & 2.70  & 3.85  &       & 1.09  & 2.08  & 3.16  &       & 1.09  & 2.16  & 3.16  &       & \textbf{0.72} & \textbf{1.82} & \textbf{2.69} &       & 0.72  & 1.67  & 2.69  &       & 0.72  & 2.07  & 3.15 \\
          & 0.30  & 0.69  & 2.18  & 5.78  &       & 0.69  & 1.66  & 3.20  &       & 0.69  & 1.58  & 3.20  &       & \textbf{0.68} & \textbf{1.53} & \textbf{3.19} &       & 0.00  & 1.32  & 2.72  &       & 0.68  & 1.61  & 3.40 \\
          & 0.50  & \textbf{0.00} & 1.38  & 2.45  &       & \textbf{0.00} & 0.90  & 1.90  &       & \textbf{0.00} & 0.91  & 1.77  &       & \textbf{0.00} & \textbf{0.71} & \textbf{1.60} &       & 0.00  & 0.64  & 1.41  &       & 0.00  & 0.84  & 1.84 \\
          & 0.70  & \textbf{0.00} & 0.63  & 2.68  &       & \textbf{0.00} & 0.39  & 1.67  &       & \textbf{0.00} & 0.39  & 1.67  &       & \textbf{0.00} & \textbf{0.33} & \textbf{1.27} &       & 0.00  & 0.18  & 1.00  &       & 0.00  & 0.38  & 1.67 \\
          & 0.90  & \textbf{0.00} & 0.35  & 2.09  &       & \textbf{0.00} & 0.35  & 2.09  &       & \textbf{0.00} & 0.35  & 2.09  &       & \textbf{0.00} & \textbf{0.34} & \textbf{2.08} &       & 0.00  & 0.26  & 2.03  &       & 0.00  & 0.37  & 2.09 \\
    0.30  & 0.10  & 1.21  & 3.11  & 5.24  &       & 1.21  & 2.48  & 4.55  &       & 1.21  & 2.44  & 4.55  &       & \textbf{0.89} & \textbf{2.31} & \textbf{4.19} &       & 0.68  & 2.17  & 3.84  &       & 1.71  & 2.55  & 4.54 \\
          & 0.30  & 1.12  & 3.37  & 6.04  &       & 1.12  & 3.12  & \textbf{4.62} &       & 2.15  & 3.46  & 6.04  &       & \textbf{1.11} & \textbf{2.94} & 5.06  &       & 1.11  & 2.55  & 4.33  &       & 1.11  & 3.18  & 5.28 \\
          & 0.50  & 1.12  & 2.40  & 5.24  &       & 1.12  & \textbf{1.52} & 3.50  &       & 2.15  & 1.85  & 3.50  &       & \textbf{0.86} & 1.65  & \textbf{3.49} &       & 0.66  & 1.48  & 3.49  &       & 0.86  & 1.87  & 3.49 \\
          & 0.70  & 0.88  & 1.97  & 5.32  &       & \textbf{0.00} & \textbf{1.61} & \textbf{4.61} &       & \textbf{0.00} & 1.90  & 5.32  &       & \textbf{0.00} & 1.69  & 4.66  &       & 0.00  & 1.32  & 4.10  &       & 0.00  & 1.94  & 5.31 \\
          & 0.90  & \textbf{0.00} & 1.27  & 3.38  &       & \textbf{0.00} & \textbf{1.05} & 2.82  &       & \textbf{0.00} & 1.10  & 2.82  &       & \textbf{0.00} & 1.06  & \textbf{2.79} &       & 0.00  & 0.84  & 2.73  &       & 0.00  & 1.23  & 2.82 \\
    0.50  & 0.10  & 2.80  & 4.88  & 6.76  &       & \textbf{1.68} & 4.17  & 6.76  &       & \textbf{1.68} & 4.11  & 6.76  &       & \textbf{1.68} & \textbf{3.87} & \textbf{6.14} &       & 1.60  & 3.59  & 5.61  &       & 1.68  & 4.38  & 6.75 \\
          & 0.30  & 2.85  & 4.49  & 7.29  &       & 1.82  & 4.22  & 7.29  &       & 1.82  & 4.12  & 7.29  &       & \textbf{1.79} & \textbf{3.99} & \textbf{7.16} &       & 1.68  & 3.85  & 7.08  &       & 1.81  & 4.25  & 7.29 \\
          & 0.50  & 1.94  & 4.48  & \textbf{6.02} &       & \textbf{0.97} & 4.28  & 6.08  &       & \textbf{0.97} & \textbf{4.25} & 6.08  &       & \textbf{0.97} & 4.32  & 6.31  &       & 0.97  & 4.08  & 6.01  &       & 0.97  & 4.67  & 7.14 \\
          & 0.70  & 0.64  & 4.20  & \textbf{7.17} &       & 0.64  & \textbf{4.16} & \textbf{7.17} &       & 0.64  & 4.19  & \textbf{7.17} &       & \textbf{0.63} & 4.26  & 7.29  &       & 0.63  & 4.16  & 7.17  &       & 0.63  & 4.74  & 8.36 \\
          & 0.90  & 1.03  & 3.37  & 7.00  &       & 1.03  & \textbf{3.05} & 7.00  &       & 1.03  & 3.13  & \textbf{6.79} &       & \textbf{0.85} & 3.10  & 7.00  &       & 0.68  & 2.97  & 7.00  &       & 1.02  & 3.42  & 7.00 \\
    0.70  & 0.10  & 0.93  & 7.72  & 15.88 &       & 0.93  & \textbf{7.43} & \textbf{14.86} &       & 0.93  & 7.53  & 16.22 &       & \textbf{0.92} & 7.46  & 15.81 &       & 0.92  & 7.35  & 14.86 &       & 0.92  & 7.56  & 16.55 \\
          & 0.30  & \textbf{5.05} & 8.64  & 12.40 &       & 5.78  & 8.49  & 12.40 &       & 5.78  & 8.53  & 12.50 &       & 5.34  & \textbf{8.43} & \textbf{12.39} &       & 5.05  & 8.33  & 12.39 &       & 5.77  & 8.61  & 13.30 \\
          & 0.50  & \textbf{5.37} & 10.45 & 14.03 &       & \textbf{5.37} & \textbf{10.05} & 14.03 &       & \textbf{5.37} & 10.17 & 14.03 &       & 5.41  & 10.11 & \textbf{14.02} &       & 5.37  & 10.01 & 14.02 &       & 5.78  & 10.55 & 14.14 \\
          & 0.70  & 1.69  & 6.90  & \textbf{12.08} &       & \textbf{1.58} & 6.56  & \textbf{12.08} &       & \textbf{1.58} & \textbf{6.47} & \textbf{12.08} &       & 1.69  & 6.80  & \textbf{12.08} &       & 1.69  & 6.69  & 12.08 &       & 1.69  & 7.18  & 12.08 \\
          & 0.90  & \textbf{0.01} & \textbf{8.12} & 13.64 &       & \textbf{0.01} & \textbf{8.12} & 13.64 &       & \textbf{0.01} & 8.27  & 13.64 &       & 0.65  & 8.21  & \textbf{13.63} &       & 0.00  & 8.06  & 13.63 &       & 1.19  & 8.51  & 13.63 \\
    0.90  & 0.10  & \textbf{0.00} & 0.59  & 4.93  &       & \textbf{0.00} & \textbf{0.49} & 4.93  &       & \textbf{0.00} & \textbf{0.49} & 4.93  &       & \textbf{0.00} & \textbf{0.49} & \textbf{4.92} &       & 0.00  & 0.49  & 4.92  &       & 0.00  & 0.49  & 4.92 \\
          & 0.30  & \textbf{0.00} & 0.16  & 1.48  &       & \textbf{0.00} & \textbf{0.00} & \textbf{0.01} &       & \textbf{0.00} & \textbf{0.00} & \textbf{0.01} &       & \textbf{0.00} & \textbf{0.00} & \textbf{0.01} &       & 0.00  & 0.00  & 0.00  &       & 0.00  & 0.01  & 0.10 \\
          & 0.50  & \textbf{0.00} & 2.58  & \textbf{12.30} &       & \textbf{0.00} & \textbf{2.49} & \textbf{12.30} &       & \textbf{0.00} & \textbf{2.49} & \textbf{12.30} &       & \textbf{0.00} & 2.56  & \textbf{12.30} &       & 0.00  & 2.49  & 12.30 &       & 0.00  & 2.72  & 12.30 \\
          & 0.70  & \textbf{0.00} & 6.81  & \textbf{20.99} &       & \textbf{0.00} & 6.67  & \textbf{20.99} &       & \textbf{0.00} & 6.67  & \textbf{20.99} &       & \textbf{0.00} & \textbf{6.64} & \textbf{20.99} &       & 0.00  & 6.61  & 20.99 &       & 0.00  & 6.75  & 20.99 \\
          & 0.90  & \textbf{0.00} & 6.09  & 19.33 &       & \textbf{0.00} & \textbf{5.99} & \textbf{19.10} &       & \textbf{0.00} & \textbf{5.99} & \textbf{19.10} &       & \textbf{0.00} & 6.02  & 19.16 &       & 0.00  & 5.99  & 19.09 &       & 0.00  & 6.05  & 19.33 \\
    Avg.  &       & 1.15  & 3.95  & 8.13  &       & 1.00  & 3.65  & 7.63  &       & 1.08  & 3.70  & 7.76  &       & \textbf{0.97} & \textbf{3.63} & \textbf{7.61} &       & 0.87  & 3.48  & 7.42  &       & 1.06  & 3.84  & 7.90 \\
    \bottomrule
    \end{tabular}%
  \label{tab4}%
\end{sidewaystable}%

\begin{sidewaystable}[htbp]
\tiny
  \centering
  \caption{Gaps (\%) of various algorithms on instances with 50 orders (single run for each instance)}
    \begin{tabular}{lllllllllllllllllllllllll}
    \toprule
    \multicolumn{2}{l}{n=50} & \multicolumn{3}{l}{ABC} &       & \multicolumn{3}{l}{HSSGA} &       & \multicolumn{3}{l}{EA/G-LS} &       & \multicolumn{3}{l}{Sparrow-Average} &       & \multicolumn{3}{l}{Sparrow-Best} &       & \multicolumn{3}{l}{Sparrow-Worst} \\
\cmidrule{1-5}\cmidrule{7-9}\cmidrule{11-13}\cmidrule{15-17}\cmidrule{19-21}\cmidrule{23-25}    $\tau$ & R     & Min   & Avg   & Max   &       & Min   & Avg   & Max   &       & Min   & Avg   & Max   &       & Min   & Avg   & Max   &       & Min   & Avg   & Max   &       & Min   & Avg   & Max \\
    \midrule
    0.10  & 0.10  & 1.20  & 1.89  & 2.81  &       & \textbf{0.51} & 1.18  & 1.69  &       & \textbf{0.51} & 1.08  & 1.47  &       & \textbf{0.51} & \textbf{0.79} & \textbf{1.27} &       & 0.37  & 0.64  & 0.99  &       & 0.51  & 0.94  & 1.59 \\
          & 0.30  & 1.06  & 1.77  & 2.33  &       & 0.71  & 1.12  & 1.57  &       & 0.71  & 1.10  & 1.57  &       & \textbf{0.60} & \textbf{1.00} & \textbf{1.53} &       & 0.45  & 0.90  & 1.53  &       & 0.70  & 1.08  & 1.53 \\
          & 0.50  & 0.34  & 0.96  & 2.03  &       & \textbf{0.00} & 0.43  & \textbf{0.84} &       & \textbf{0.00} & 0.45  & 0.95  &       & \textbf{0.00} & \textbf{0.42} & 0.96  &       & 0.00  & 0.33  & 0.83  &       & 0.00  & 0.56  & 1.25 \\
          & 0.70  & \textbf{0.00} & 1.90  & 16.39 &       & \textbf{0.00} & 1.73  & 16.39 &       & \textbf{0.00} & \textbf{1.66} & 16.39 &       & \textbf{0.00} & \textbf{1.66} & \textbf{16.38} &       & 0.00  & 1.63  & 16.38 &       & 0.00  & 1.78  & 16.38 \\
          & 0.90  & \textbf{0.00} & \textbf{0.00} & \textbf{0.00} &       & \textbf{0.00} & \textbf{0.00} & \textbf{0.00} &       & \textbf{0.00} & \textbf{0.00} & \textbf{0.00} &       & \textbf{0.00} & \textbf{0.00} & \textbf{0.00} &       & 0.00  & 0.00  & 0.00  &       & 0.00  & 0.00  & 0.00 \\
    0.30  & 0.10  & 1.63  & 2.59  & 3.77  &       & 1.02  & 1.80  & \textbf{2.79} &       & 1.22  & 1.93  & 2.86  &       & \textbf{0.63} & \textbf{1.59} & 2.86  &       & 0.60  & 1.30  & 2.48  &       & 0.81  & 1.85  & 3.24 \\
          & 0.30  & 1.94  & 2.98  & 4.49  &       & 1.55  & 2.06  & \textbf{2.66} &       & 1.54  & 2.18  & 3.05  &       & \textbf{1.40} & \textbf{1.81} & 2.69  &       & 1.15  & 1.63  & 2.51  &       & 1.54  & 2.13  & 3.23 \\
          & 0.50  & 1.94  & 1.88  & 3.88  &       & 1.55  & 1.51  & 3.88  &       & 1.54  & 1.53  & 3.68  &       & \textbf{0.18} & \textbf{1.44} & \textbf{3.64} &       & 0.00  & 1.23  & 3.10  &       & 0.37  & 1.63  & 4.26 \\
          & 0.70  & \textbf{0.00} & 0.75  & 1.56  &       & \textbf{0.00} & 0.39  & \textbf{1.03} &       & \textbf{0.00} & \textbf{0.35} & 1.04  &       & \textbf{0.00} & 0.39  & 1.16  &       & 0.00  & 0.24  & 1.03  &       & 0.00  & 0.66  & 1.55 \\
          & 0.90  & \textbf{0.00} & 0.50  & 1.83  &       & \textbf{0.00} & 0.29  & 1.21  &       & \textbf{0.00} & \textbf{0.23} & \textbf{1.02} &       & \textbf{0.00} & 0.27  & 1.32  &       & 0.00  & 0.16  & 1.01  &       & 0.00  & 0.35  & 1.62 \\
    0.50  & 0.10  & 1.71  & 3.07  & 4.17  &       & \textbf{1.11} & 2.20  & 3.20  &       & \textbf{1.11} & 2.23  & 3.38  &       & \textbf{1.11} & \textbf{2.01} & \textbf{2.64} &       & 0.76  & 1.78  & 2.43  &       & 1.11  & 2.15  & 3.05 \\
          & 0.30  & 3.08  & 4.45  & 6.11  &       & 2.26  & 3.43  & \textbf{5.15} &       & 2.26  & 3.53  & 5.73  &       & \textbf{1.97} & \textbf{3.30} & 5.44  &       & 1.64  & 3.01  & 5.15  &       & 2.25  & 3.78  & 5.72 \\
          & 0.50  & 2.07  & 3.78  & 6.82  &       & \textbf{1.03} & 3.04  & \textbf{6.33} &       & \textbf{1.03} & 3.17  & 7.14  &       & \textbf{1.03} & \textbf{3.00} & 6.49  &       & 0.83  & 2.77  & 6.16  &       & 1.03  & 3.39  & 7.14 \\
          & 0.70  & 0.76  & 2.28  & 4.24  &       & \textbf{0.36} & 2.00  & \textbf{3.87} &       & 0.38  & 2.03  & 4.24  &       & 0.37  & \textbf{1.97} & 4.14  &       & 0.36  & 1.67  & 3.86  &       & 0.37  & 2.22  & 4.41 \\
          & 0.90  & 3.61  & 1.71  & \textbf{4.04} &       & 0.00  & 1.54  & 4.31  &       & -4.02 & 1.57  & \textbf{4.04} &       & \textbf{-4.09} & \textbf{1.53} & \textbf{4.04} &       & -4.41 & 1.17  & 4.04  &       & -4.01 & 1.91  & 4.04 \\
    0.70  & 0.10  & 3.16  & 4.57  & 5.79  &       & 2.42  & 4.17  & 5.20  &       & 2.42  & 4.18  & 5.39  &       & \textbf{2.24} & \textbf{3.88} & \textbf{4.74} &       & 1.85  & 3.58  & 4.45  &       & 2.41  & 4.13  & 5.19 \\
          & 0.30  & 3.51  & 5.74  & 9.36  &       & 3.01  & 5.02  & \textbf{8.30} &       & 3.17  & 5.05  & \textbf{8.30} &       & \textbf{2.75} & \textbf{4.82} & 8.40  &       & 2.47  & 4.46  & 8.12  &       & 3.00  & 5.12  & 9.18 \\
          & 0.50  & 5.30  & 7.09  & 11.39 &       & 5.08  & 6.71  & 11.03 &       & 5.17  & 6.70  & 11.03 &       & \textbf{4.74} & \textbf{6.46} & \textbf{10.91} &       & 4.42  & 6.10  & 10.32 &       & 5.08  & 6.88  & 11.28 \\
          & 0.70  & 1.89  & 7.56  & 15.26 &       & 1.72  & 7.18  & 15.38 &       & \textbf{1.37} & \textbf{7.09} & \textbf{14.31} &       & 1.73  & 7.13  & 15.50 &       & 1.37  & 6.69  & 15.38 &       & 2.40  & 7.63  & 15.81 \\
          & 0.90  & 3.69  & 8.02  & 13.67 &       & \textbf{2.91} & 7.69  & 13.38 &       & 3.30  & \textbf{7.62} & \textbf{13.00} &       & 3.26  & 7.63  & 13.02 &       & 2.91  & 7.28  & 13.00 &       & 3.68  & 7.95  & 13.10 \\
    0.90  & 0.10  & 7.30  & 11.33 & \textbf{16.77} &       & \textbf{6.55} & \textbf{10.97} & \textbf{16.77} &       & 7.30  & 11.15 & \textbf{16.77} &       & \textbf{6.55} & 10.98 & \textbf{16.77} &       & 6.55  & 10.87 & 16.77 &       & 6.55  & 11.23 & 16.77 \\
          & 0.30  & \textbf{9.28} & 13.77 & 17.80 &       & \textbf{9.28} & 13.49 & 17.60 &       & \textbf{9.28} & 13.47 & \textbf{17.40} &       & \textbf{9.28} & \textbf{13.43} & 17.80 &       & 9.28  & 13.27 & 17.60 &       & 9.28  & 13.57 & 18.00 \\
          & 0.50  & 3.28  & 12.77 & 17.29 &       & 3.28  & 12.33 & 16.88 &       & 3.46  & 12.45 & \textbf{16.73} &       & \textbf{3.05} & \textbf{12.31} & 16.78 &       & 2.62  & 12.16 & 16.51 &       & 3.64  & 12.67 & 17.65 \\
          & 0.70  & 5.81  & 11.53 & \textbf{17.85} &       & \textbf{5.62} & 11.48 & \textbf{17.85} &       & \textbf{5.62} & \textbf{11.39} & \textbf{17.85} &       & \textbf{5.62} & 11.50 & 18.42 &       & 5.62  & 11.26 & 17.84 &       & 5.62  & 11.74 & 19.03 \\
          & 0.90  & 10.02 & 12.38 & 16.27 &       & \textbf{9.57} & 12.37 & \textbf{15.92} &       & 9.58  & \textbf{12.30} & 16.09 &       & 9.74  & 12.49 & 16.30 &       & 9.40  & 12.22 & 16.10 &       & 10.01 & 12.81 & 16.46 \\
    Avg.  &       & 2.90  & 5.01  & 8.24  &       & 2.38  & 4.57  & 7.73  &       & 2.28  & 4.58  & 7.74  &       & \textbf{2.11} & \textbf{4.47} & \textbf{7.73} &       & 1.93  & 4.25  & 7.50  &       & 2.25  & 4.73  & 8.06 \\
    \bottomrule
    \end{tabular}%
  \label{tab5}%
\end{sidewaystable}%

\begin{sidewaystable}[htbp]
\tiny
  \centering
  \caption{Gaps (\%) of various algorithms on instances with 100 orders (single run for each instance)}
    \begin{tabular}{lllllllllllllllllllllllll}
    \toprule
    \multicolumn{2}{l}{n=100} & \multicolumn{3}{l}{ABC} &       & \multicolumn{3}{l}{HSSGA} &       & \multicolumn{3}{l}{EA/G-LS} &       & \multicolumn{3}{l}{Sparrow-Average} &       & \multicolumn{3}{l}{Sparrow-Best} &       & \multicolumn{3}{l}{Sparrow-Worst} \\
\cmidrule{1-5}\cmidrule{7-9}\cmidrule{11-13}\cmidrule{15-17}\cmidrule{19-21}\cmidrule{23-25}    $\tau$ & R     & Min   & Avg   & Max   &       & Min   & Avg   & Max   &       & Min   & Avg   & Max   &       & Min   & Avg   & Max   &       & Min   & Avg   & Max   &       & Min   & Avg   & Max \\
    \midrule
    0.10  & 0.10  & 1.11  & 1.75  & 2.21  &       & 0.40  & 0.77  & 1.32  &       & 0.72  & 0.94  & 1.22  &       & \textbf{0.39} & \textbf{0.57} & \textbf{0.80} &       & 0.19  & 0.46  & 0.65  &       & 0.54  & 0.69  & 0.85 \\
          & 0.30  & 0.90  & 1.35  & 1.71  &       & 0.45  & 0.71  & 0.98  &       & 0.44  & 0.74  & 1.06  &       & \textbf{0.34} & \textbf{0.61} & \textbf{0.88} &       & 0.18  & 0.50  & 0.81  &       & 0.45  & 0.75  & 1.01 \\
          & 0.50  & 0.36  & 0.70  & 1.23  &       & \textbf{0.00} & 0.19  & 0.38  &       & \textbf{0.00} & 0.20  & 0.37  &       & \textbf{0.00} & \textbf{0.13} & \textbf{0.28} &       & 0.00  & 0.07  & 0.18  &       & 0.00  & 0.20  & 0.37 \\
          & 0.70  & \textbf{0.00} & 0.09  & 0.35  &       & \textbf{0.00} & \textbf{0.00} & \textbf{0.00} &       & \textbf{0.00} & \textbf{0.00} & \textbf{0.00} &       & \textbf{0.00} & \textbf{0.00} & \textbf{0.00} &       & 0.00  & 0.00  & 0.00  &       & 0.00  & 0.00  & 0.00 \\
          & 0.90  & \textbf{0.00} & \textbf{0.00} & \textbf{0.00} &       & \textbf{0.00} & \textbf{0.00} & \textbf{0.00} &       & \textbf{0.00} & \textbf{0.00} & \textbf{0.00} &       & \textbf{0.00} & \textbf{0.00} & \textbf{0.00} &       & 0.00  & 0.00  & 0.00  &       & 0.00  & 0.00  & 0.00 \\
    0.30  & 0.10  & 1.27  & 2.11  & 2.71  &       & \textbf{0.50} & 1.18  & 1.71  &       & 0.69  & 1.25  & 1.71  &       & 0.53  & \textbf{0.99} & \textbf{1.44} &       & 0.39  & 0.83  & 1.35  &       & 0.58  & 1.09  & 1.53 \\
          & 0.30  & 1.40  & 2.12  & 3.97  &       & 0.66  & 1.27  & 2.64  &       & 0.66  & 1.41  & 2.64  &       & \textbf{0.57} & \textbf{1.24} & \textbf{2.49} &       & 0.56  & 1.06  & 2.23  &       & 0.74  & 1.38  & 2.64 \\
          & 0.50  & 1.40  & 1.67  & 2.45  &       & 0.66  & \textbf{1.07} & \textbf{1.52} &       & 0.66  & 1.11  & 1.69  &       & \textbf{0.65} & \textbf{1.07} & 1.75  &       & 0.45  & 0.91  & 1.52  &       & 0.81  & 1.22  & 2.03 \\
          & 0.70  & \textbf{0.00} & 0.63  & 1.16  &       & \textbf{0.00} & 0.35  & 0.85  &       & \textbf{0.00} & \textbf{0.24} & \textbf{0.63} &       & \textbf{0.00} & 0.30  & 0.79  &       & 0.00  & 0.17  & 0.62  &       & 0.00  & 0.38  & 0.96 \\
          & 0.90  & \textbf{0.00} & 0.26  & 0.82  &       & \textbf{0.00} & \textbf{0.10} & 0.56  &       & \textbf{0.00} & 0.14  & 0.47  &       & \textbf{0.00} & 0.11  & \textbf{0.42} &       & 0.00  & 0.05  & 0.28  &       & 0.00  & 0.20  & 0.64 \\
    0.50  & 0.10  & 2.50  & 3.37  & 4.33  &       & 1.22  & 2.09  & 2.99  &       & 1.69  & 2.31  & 3.27  &       & \textbf{1.07} & \textbf{1.79} & \textbf{2.71} &       & 0.81  & 1.57  & 2.47  &       & 1.26  & 2.03  & 2.91 \\
          & 0.30  & 2.46  & 3.01  & 4.00  &       & 1.54  & 2.29  & 2.95  &       & 1.67  & 2.43  & 2.96  &       & \textbf{1.49} & \textbf{2.16} & \textbf{2.63} &       & 1.14  & 1.94  & 2.51  &       & 1.53  & 2.45  & 3.15 \\
          & 0.50  & 2.26  & 3.13  & 4.21  &       & \textbf{1.22} & 2.36  & \textbf{3.35} &       & 1.48  & 2.51  & 3.61  &       & 1.31  & \textbf{2.33} & \textbf{3.35} &       & 1.13  & 2.12  & 3.17  &       & 1.39  & 2.66  & 3.69 \\
          & 0.70  & 0.77  & 1.82  & 2.99  &       & 0.51  & 1.49  & 2.33  &       & 0.51  & \textbf{1.32} & \textbf{2.24} &       & \textbf{0.35} & 1.49  & 2.38  &       & 0.34  & 1.26  & 2.06  &       & 0.51  & 1.70  & 2.78 \\
          & 0.90  & 0.53  & 1.42  & 3.13  &       & \textbf{0.18} & 1.04  & \textbf{2.15} &       & \textbf{0.18} & \textbf{0.90} & \textbf{2.15} &       & 0.19  & 0.92  & 2.40  &       & 0.00  & 0.73  & 2.23  &       & 0.35  & 1.11  & 2.77 \\
    0.70  & 0.10  & 3.20  & 3.99  & 4.84  &       & 2.09  & 2.84  & 3.53  &       & 2.27  & 2.98  & 3.82  &       & \textbf{1.69} & \textbf{2.47} & \textbf{3.21} &       & 1.45  & 2.25  & 3.13  &       & 1.90  & 2.73  & 3.52 \\
          & 0.30  & 2.37  & 4.72  & 7.16  &       & 1.71  & 3.88  & 6.07  &       & 1.67  & 4.09  & 6.26  &       & \textbf{1.35} & \textbf{3.66} & \textbf{5.80} &       & 1.14  & 3.35  & 5.43  &       & 1.66  & 3.95  & 6.07 \\
          & 0.50  & 3.27  & 5.27  & 11.94 &       & 2.25  & 4.62  & 10.95 &       & 2.62  & 4.69  & 10.95 &       & \textbf{2.23} & \textbf{4.34} & \textbf{10.65} &       & 1.97  & 4.06  & 10.44 &       & 2.44  & 4.60  & 11.02 \\
          & 0.70  & 2.77  & 5.37  & 7.37  &       & 2.18  & 5.16  & 7.78  &       & 2.77  & 5.39  & 7.65  &       & \textbf{1.92} & \textbf{4.81} & \textbf{7.33} &       & 1.76  & 4.46  & 7.00  &       & 2.35  & 5.19  & 7.74 \\
          & 0.90  & 3.57  & 5.90  & 8.33  &       & 3.73  & 5.48  & 7.49  &       & 3.15  & 5.34  & 6.88  &       & \textbf{2.97} & \textbf{5.07} & \textbf{6.55} &       & 2.75  & 4.72  & 6.16  &       & 3.26  & 5.49  & 6.95 \\
    0.90  & 0.10  & 4.70  & 6.62  & 9.44  &       & 4.57  & 6.27  & 9.06  &       & 4.75  & 6.45  & 9.34  &       & \textbf{3.44} & \textbf{5.46} & \textbf{8.47} &       & 3.26  & 5.21  & 8.28  &       & 3.63  & 5.72  & 8.76 \\
          & 0.30  & 5.39  & 9.40  & 11.86 &       & 5.11  & 9.03  & 11.25 &       & 5.39  & 9.09  & 12.35 &       & \textbf{4.90} & \textbf{8.65} & \textbf{11.20} &       & 4.60  & 8.40  & 10.96 &       & 5.29  & 8.98  & 11.82 \\
          & 0.50  & 8.82  & 11.88 & 17.25 &       & 8.60  & 11.26 & 16.29 &       & 8.28  & 11.20 & 15.84 &       & \textbf{7.90} & \textbf{10.76} & \textbf{15.67} &       & 7.52  & 10.47 & 15.52 &       & 8.32  & 11.01 & 15.86 \\
          & 0.70  & 7.96  & 11.19 & \textbf{12.85} &       & 8.07  & 11.23 & 13.07 &       & 8.05  & 11.07 & 13.06 &       & \textbf{7.38} & \textbf{10.73} & 13.02 &       & 7.10  & 10.41 & 12.54 &       & 7.71  & 11.14 & 13.43 \\
          & 0.90  & 4.11  & 11.31 & 16.83 &       & 3.57  & 11.07 & 16.60 &       & 4.07  & 11.02 & 16.73 &       & \textbf{3.32} & \textbf{10.67} & \textbf{16.53} &       & 3.17  & 10.41 & 16.17 &       & 3.47  & 10.92 & 16.93 \\
    Avg.  &       & 2.44  & 3.96  & 5.73  &       & 1.97  & 3.43  & 5.03  &       & 2.07  & 3.47  & 5.08  &       & \textbf{1.76} & \textbf{3.21} & \textbf{4.83} &       & 1.60  & 3.02  & 4.63  &       & 1.93  & 3.42  & 5.10 \\
    \bottomrule
    \end{tabular}%
  \label{tab6}%
\end{sidewaystable}%

Sparrow, TS, DRGA, GA, HH, LOS and ILS are run ten times on each of the instances. The average gaps of the ten runs for each instance are calculated first. Then for each group of instances with the same parameters, we calculate the minimum, average and maximum gaps of the ten instances in this group. The results of instances with 25--100 orders are shown in Tables \ref{tab1}--\ref{tab3}. The other three algorithms, ABC, HSSGA and EA/G-LS are run only once for each instance. In order to compare with these three algorithms, we calculate the minimum, average and maximum results of the best, the average, and the worst of the ten runs (denoted as Sparrow-Best, Sparrow-Average and Sparrow-Worst respectively) as a reference of the performance of our algorithm. The results are shown in Tables \ref{tab4}--\ref{tab6}. Note that in Tables \ref{tab4}--\ref{tab6}, we only highlight the best results among ABC, HSSGA, EA/G-LS and Sparrow-Average.

In Tables~\ref{tab1}--\ref{tab3}, TS, DRGA, GA, HH and LOS only reported rounded-down integer values. But it is still obvious that Sparrow produces the best solutions on nearly all the instances. In Tables~\ref{tab4}--\ref{tab6}, the average results of Sparrow are better than those of ABC, HSSGA and EA/G-LS on most of the instances. Even the worst results of the ten runs of Sparrow are better than those of ABC, HSSGA and EA/G-LS on around half of the instances. Therefore we can conclude that on average Sparrow produces the best solutions among all the algorithms.

Then we study how the performance gaps between Sparrow and other algorithms change with the different parameters of the instances. We evaluate the performance gap between an algorithm $A$ and Sparrow by the following formula:
\begin{equation}
   GapToSparrow_A = \frac{Gap_A - Gap_{Sparrow}}{Gap_{Sparrow}}
\end{equation}
where $Gap_A$ is the gap between the algorithm $A$ and the upper bound and $Gap_{Sparrow}$ is the gap between Sparrow and the upper bound.

\begin{figure}[htbp]
\centering
\includegraphics[height=1.5in]{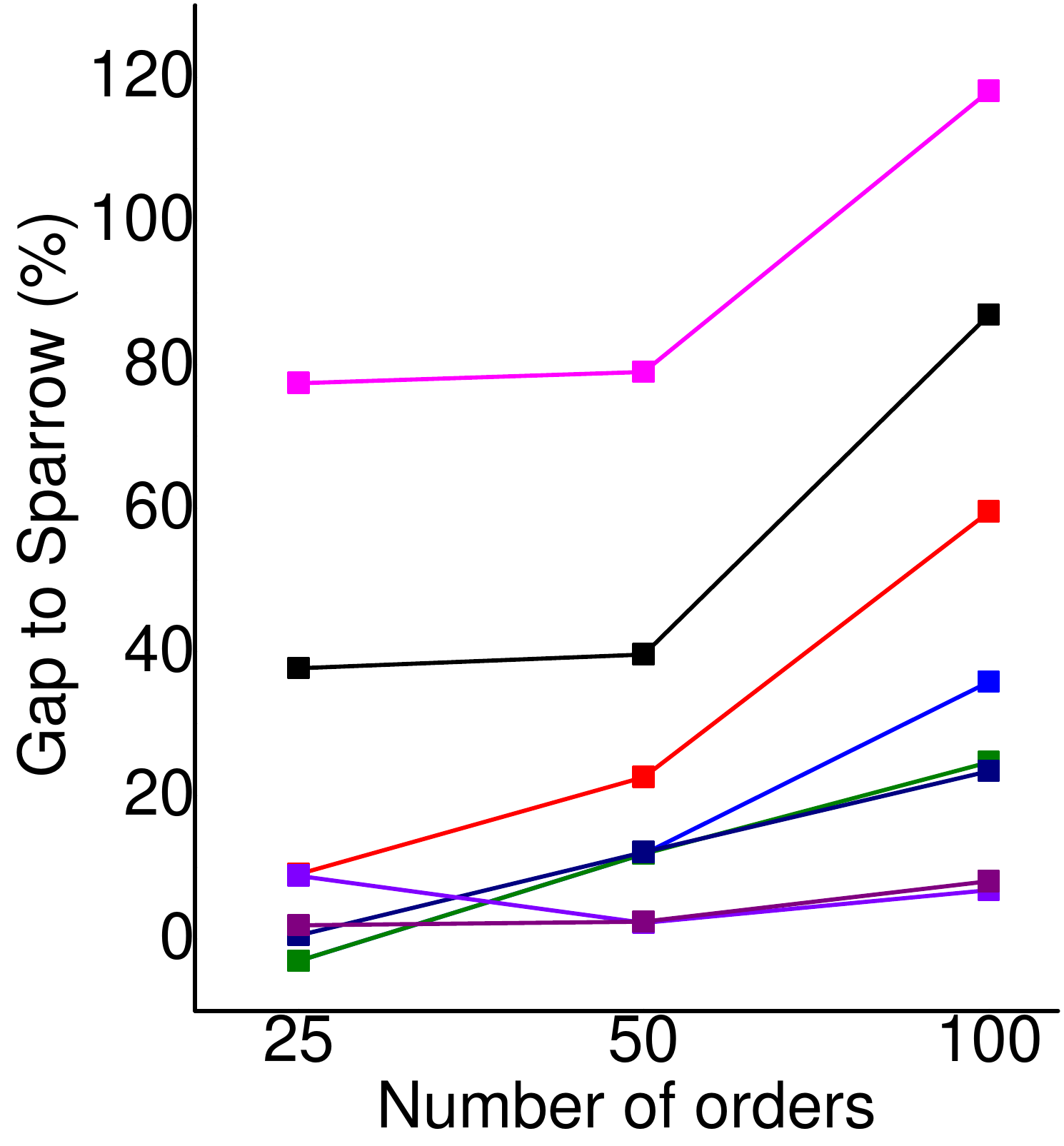}
\hfill
\includegraphics[height=1.5in]{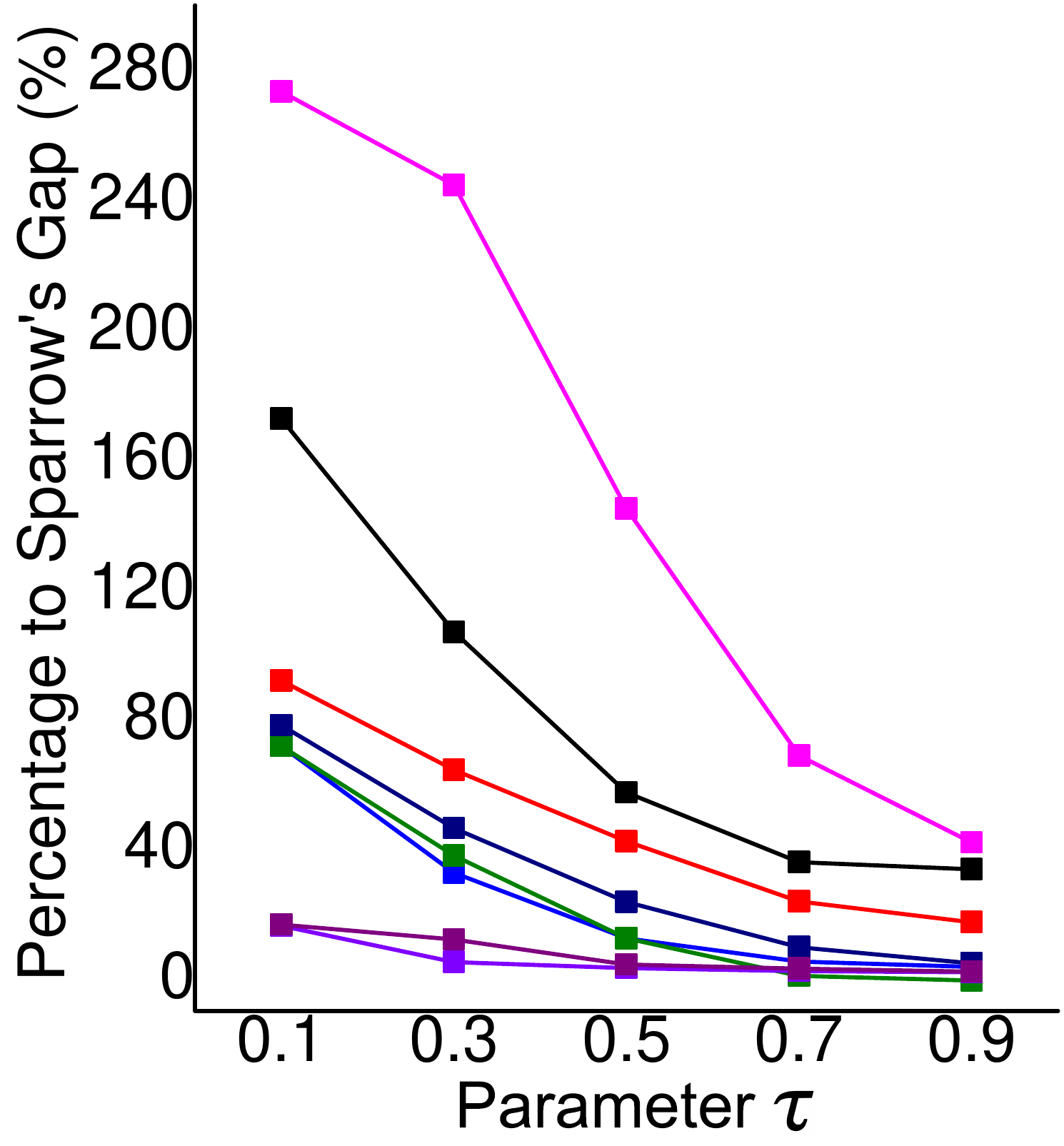}
\hfill
\includegraphics[height=1.5in]{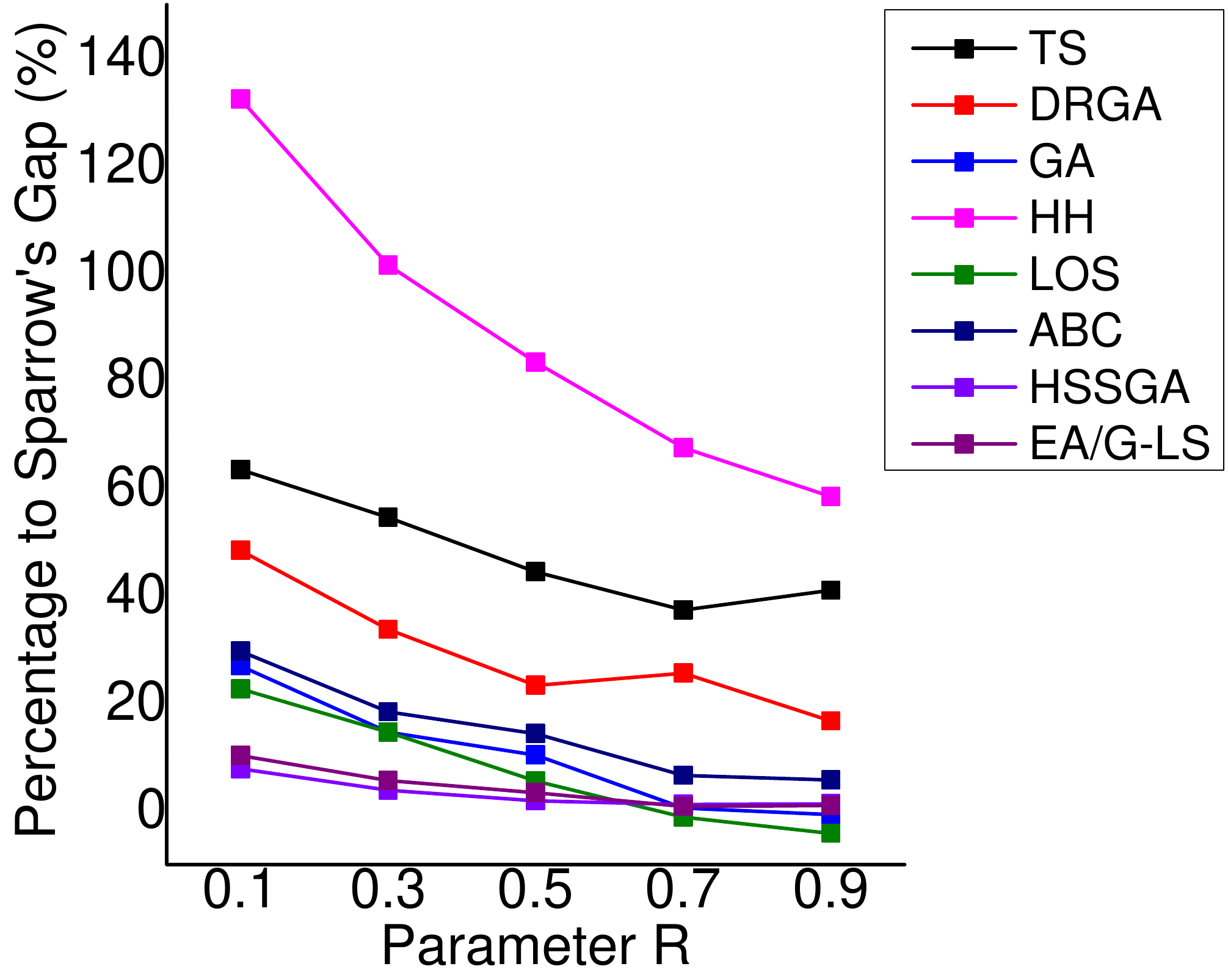}
\caption{The gaps of different algorithms to Sparrow.\protect\footnotemark}\label{figgap}
\end{figure}

\footnotetext{We do not plot the results of ILS in this figure because we find that some gaps reported by Silva et al.~\cite{silva2018exact} are smaller than the gaps between their tight upper bounds and the upper bounds by Cesaret et al.~\cite{cesaret2012tabu}.}

The performance gaps of different algorithms are calculated according to the different numbers of orders, $\tau$ and $R$ and shown in Figure~\ref{figgap}. Note that in Figure~\ref{figgap} some performance gaps of LOS are smaller than 0, because LOS only reported rounded-down integer values. In general, from Figure~\ref{figgap} we can find that gaps between the performance of Sparrow and those of other algorithms tend to grow when there are more orders, and when $\tau$ and $R$ are smaller. These are the cases found to be hard for Sparrow, but these cases are even harder for the other algorithms. 
 Among other algorithms, HSSGA and EA/G-LS have the steadiest performance for instances with different parameters. Like Sparrow, HSSGA and EA/G-LS also work better than others when there are more orders, and when $\tau$ and $R$ are smaller. On the contrary, GA and LOS work better when there are fewer orders, and when $\tau$ and $R$ are larger.

\subsection{Problem properties}\label{sec_property}

In this section, we aim to understand the problem further by studying how different properties (other than $n$, $\tau$ and $R$) of the problem correlate with its difficulty. To achieve this, we evaluate the performance of Sparrow on instances with different properties, because according to the comparison with other algorithms above, we find that Sparrow is the new state of the art for this problem. In this section, we use the same benchmark as in Section~\ref{sec_para} and \ref{sec_comparison}, and the tight upper bounds provided by Silva et al.~\cite{silva2018exact}. As in Figure~\ref{figpara}, we select the instances with 25 orders in order to decrease the influence from the upper bounds, because the upper bounds for instances with 25 orders are tighter than those for instances with 50 and 100 orders.

In this section, we calculate the following properties of each problem instance: standard deviation of setup times; standard deviation of window length; standard deviation of revenue of orders; the length of horizon; average window length; congestion ratio (this value is calculated by adding up the processing time and the smallest setup time of each order, divided by the horizon, to evaluate how congested the instance is); standard deviation of order processing time; average conflict ratio (the conflict ratio of an order is calculated by adding up the overlap time between its time window and the time windows of all other orders, divided by the length of its time window); standard deviation of conflict ratio; setup window ratio (this value is the average setup time divided by the average length of window); process window ratio (this value is the average processing time divided by the average length of window); correlation coefficient between processing time and revenue of orders.

\begin{figure}[htbp]
\centering
{\includegraphics[width=2.3in]{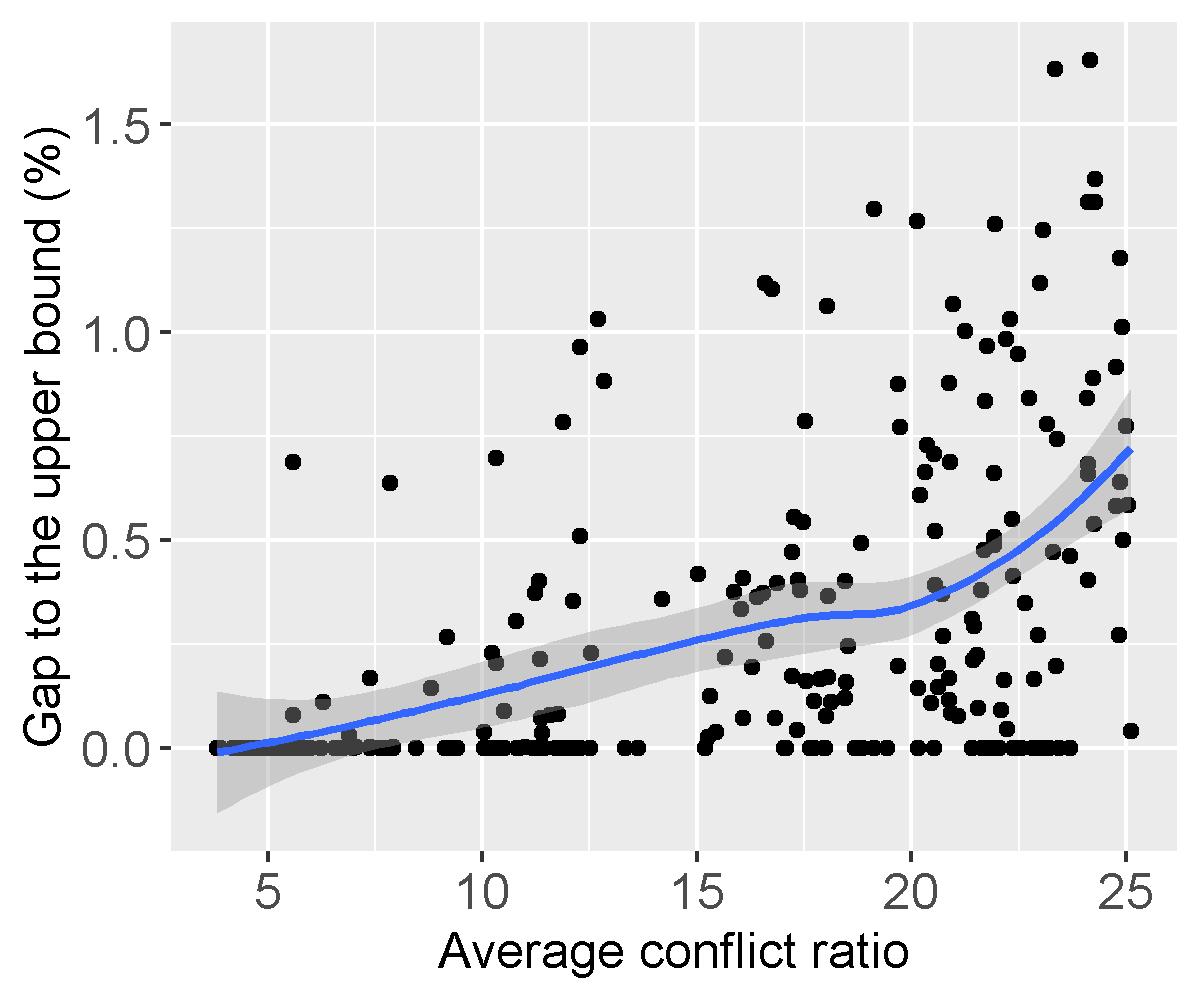}\hfill \includegraphics[width=2.3in]{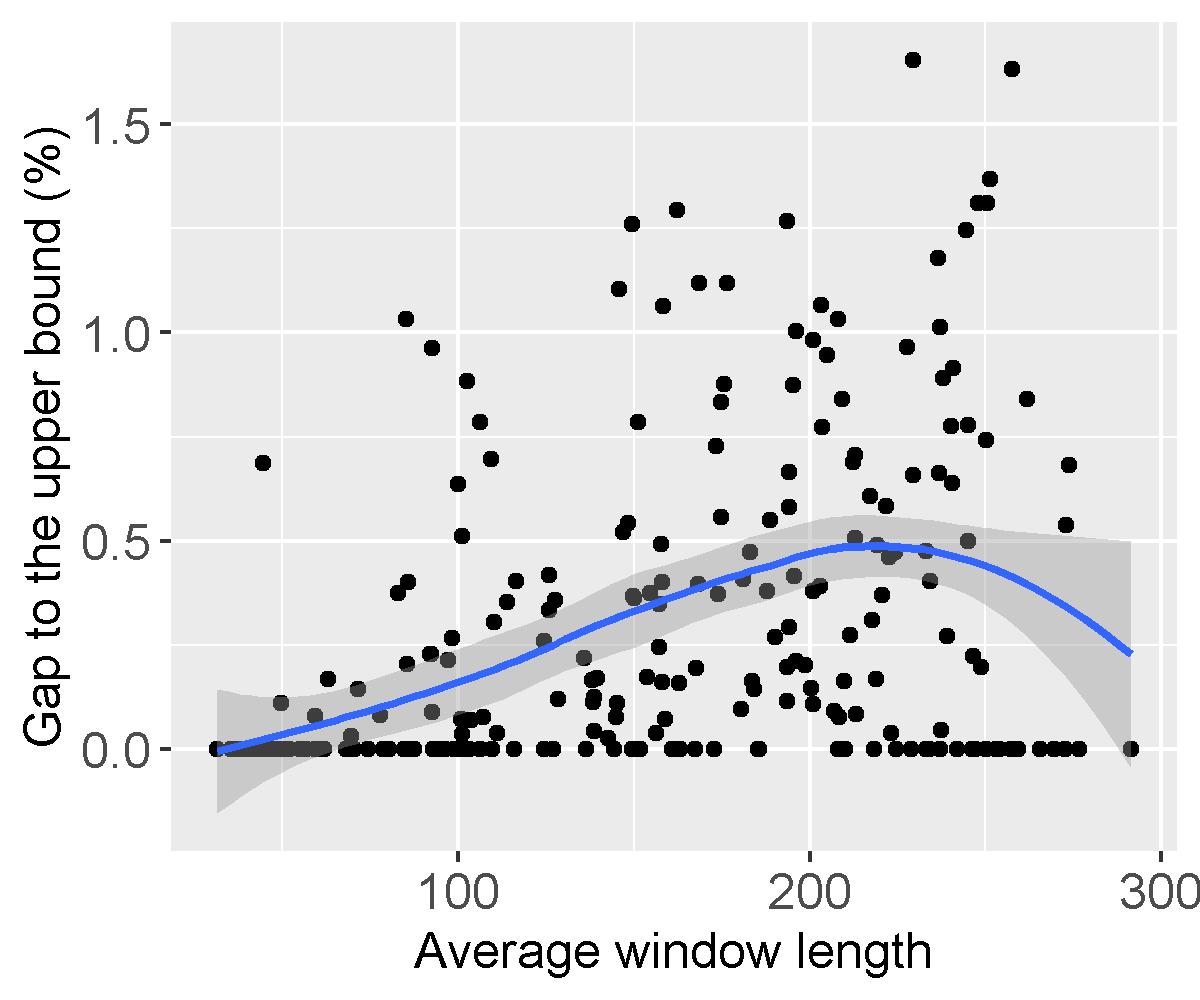}}
\vfill
{\includegraphics[width=2.3in]{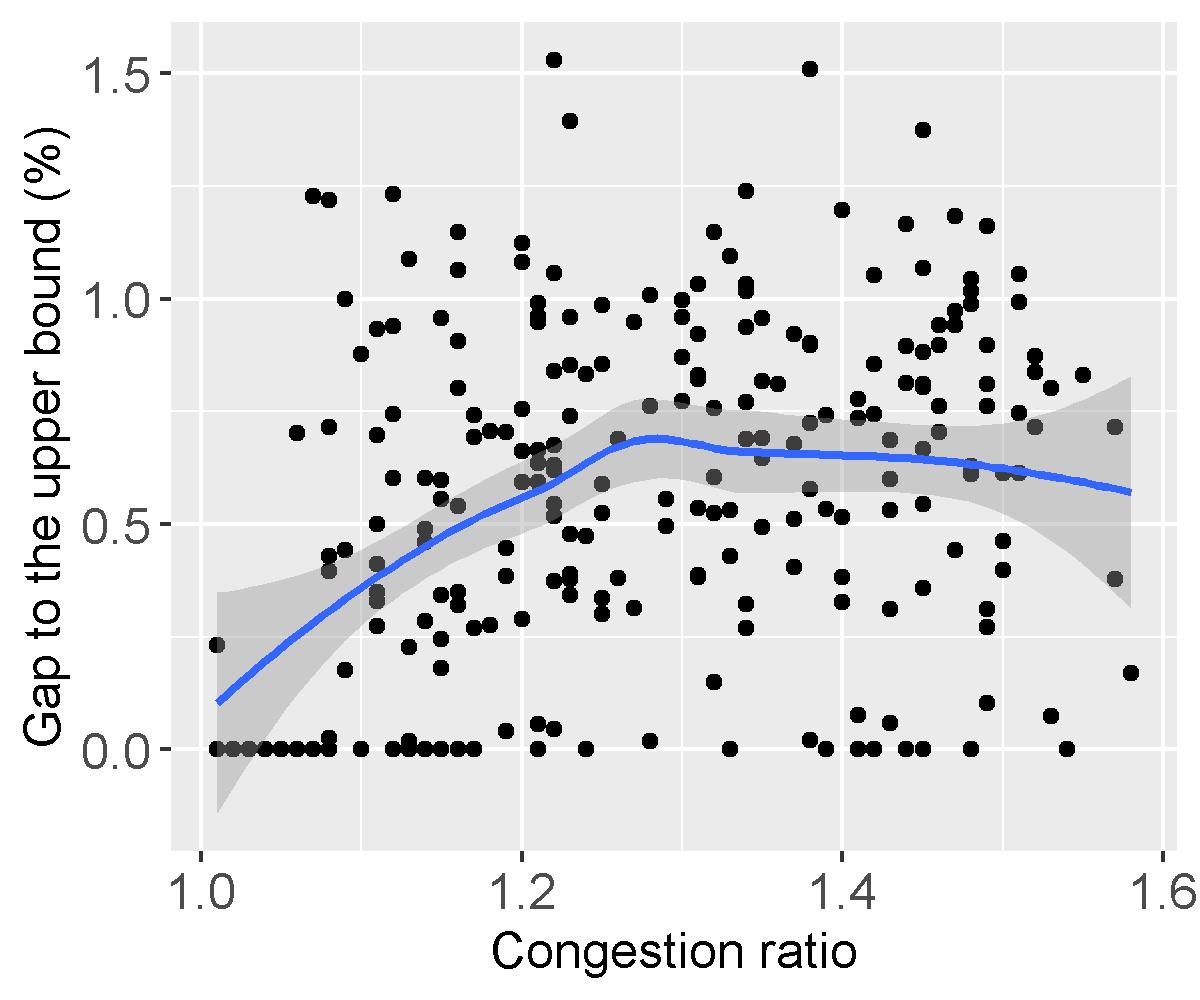}\hfill \includegraphics[width=2.3in]{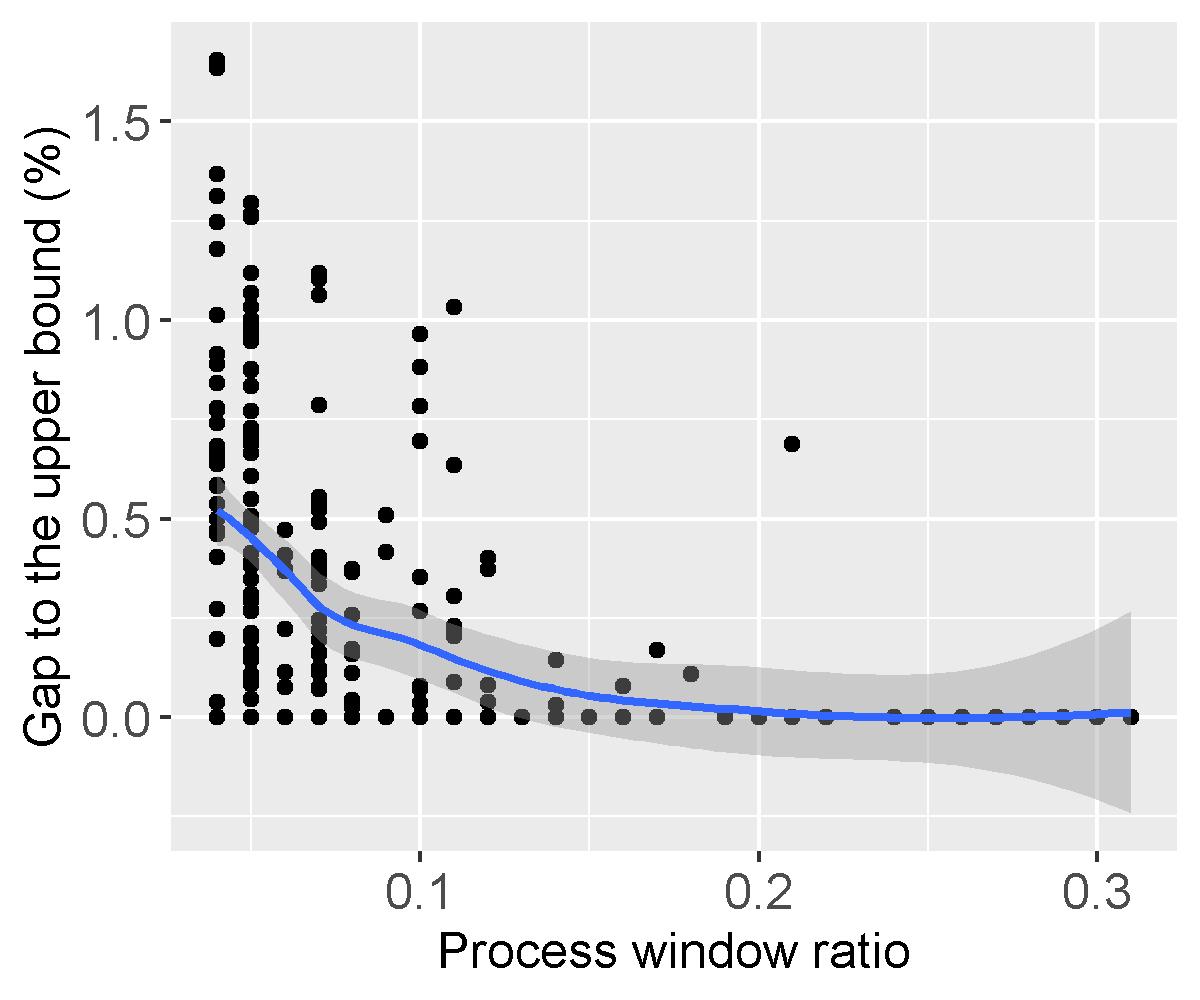}}
\vfill
{\includegraphics[width=2.3in]{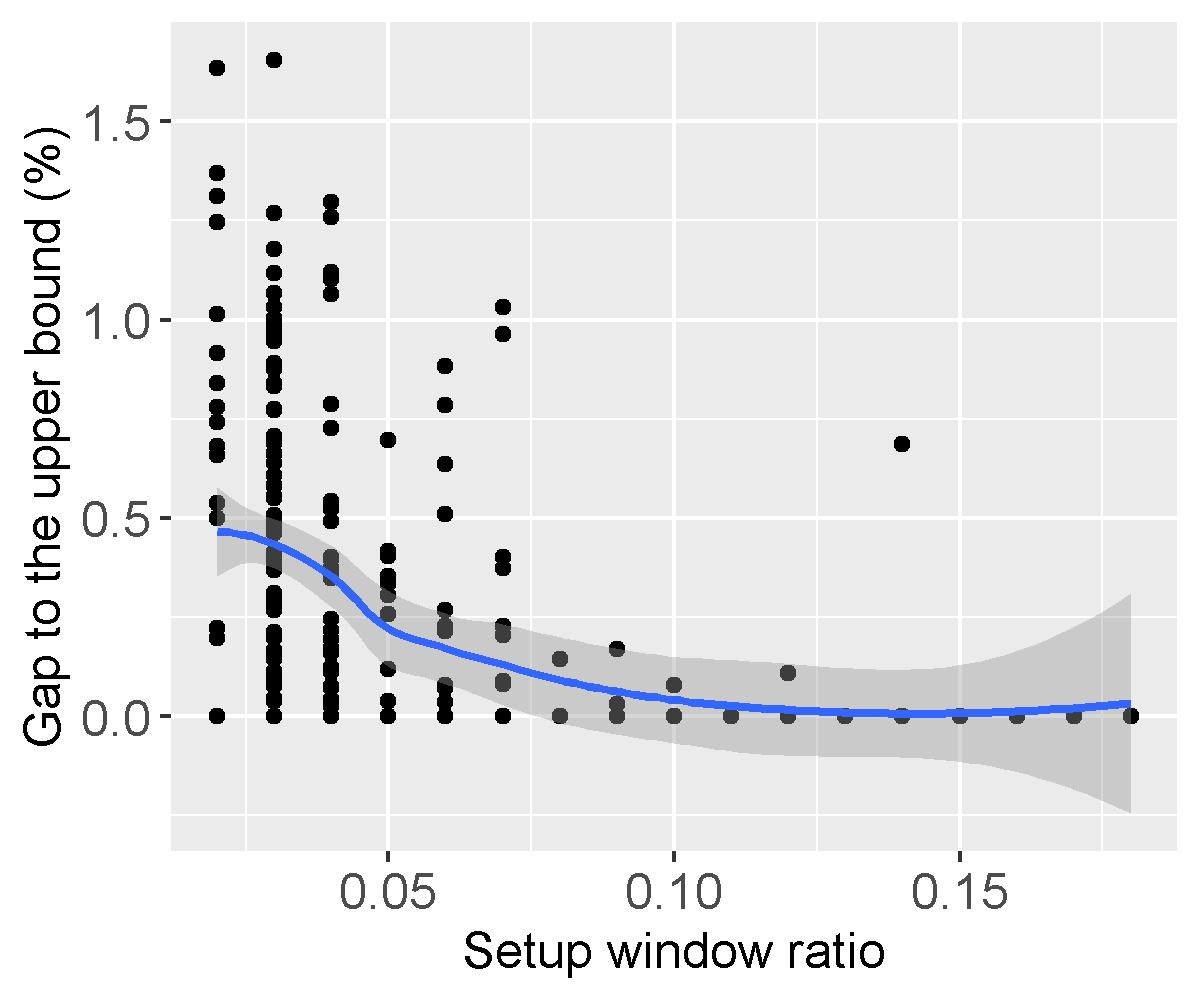}\hfill \includegraphics[width=2.3in]{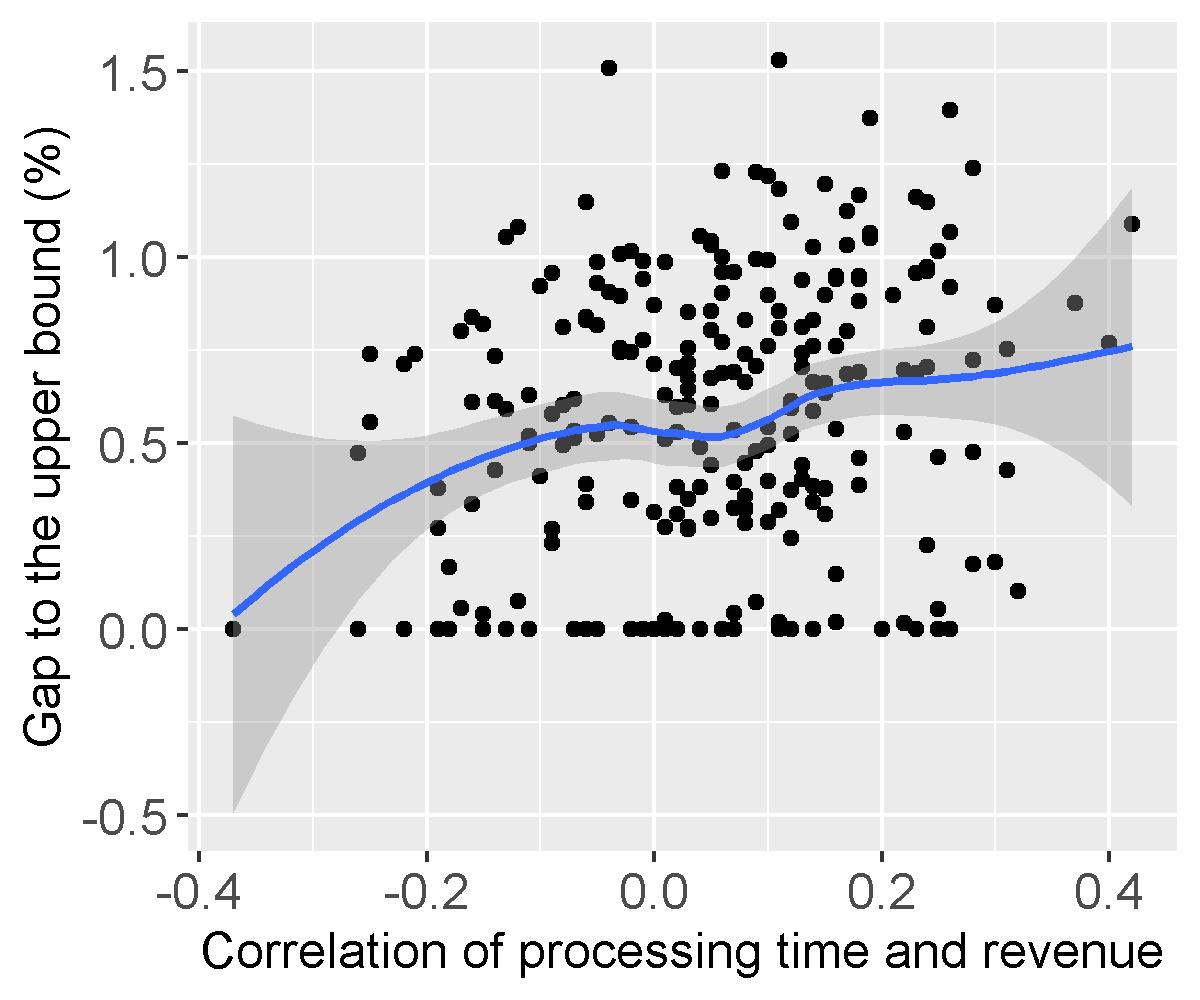}}
\caption{The correlation of the performance of Sparrow and different properties of the problem. Each point is the average of 10 runs of each instance with 25 orders.}\label{figpro}
\end{figure}

In Figure~\ref{figpro} we show how the performance of Sparrow changes with six properties. We do not plot the performance with other properties because the correlations are weak, which shows that Sparrow adapts it well for these properties. According to Figure~\ref{figpro}, the problem becomes harder for Sparrow when: (1) the average conflict ratio is larger; (2) the average window length is larger. This is consistent with the observation in Figure~\ref{figpara}, because when $\tau$ is smaller, the average window length is larger; (3) the congestion ratio is larger; (4) the process window ratio is smaller. This case is harder because the space where an order can be shifted within a window is larger. The solution space becomes larger; (5) the setup window ratio is smaller; (6) 
the correlation coefficient between processing time and revenue is larger. The reason is that when the correlation is larger, the difference among order values is smaller, therefore it is more difficult to select orders.

According to the observations above, it can be concluded that the following properties make the problem harder:

\begin{enumerate}
\item Congestion-related properties: larger congestion ratio; larger conflict ratio.

\item Order-related properties: larger correlation between processing time and revenue of orders.

\item Window-related properties: longer time windows; smaller process/setup window ratio.
\end{enumerate}

It is interesting to see that each of the three hard cases corresponds to a real-life situation. The satellite scheduling problem usually has the problem of the scheduling horizon being heavily contested; many real-life situations such as commerce cause longer orders to be more expensive, where the processing time and revenue of an order are more correlated; cases such as the travelling repairman problem involve mostly short processing times, while having relatively longer time windows.

Although the benchmark set by Cesaret et al.~\cite{cesaret2012tabu} is known to contain difficult instances, the three real-life and difficult scenarios above do not correspond to the instances generated with this methodology. Therefore in the next section, we generate three new sets of instances, each corresponding to one of the scenarios, to compare the performances of different algorithms.

\subsection{The performances of different algorithms on new instances}

In this section, we describe an evaluation of the state-of-the-art algorithms on three new sets of instances: the \emph{satellite scheduling} set is more congested by setting the parameters as follows: $n=\{100,150,200,250,300\}$, $\tau=0.1$, and $R=0.1$. The \emph{commerce} set has a more correlated processing time and revenue by generating a random value $\gamma$ in [1,20] and calculating the final revenue of order $o_i$ by $((1-q)\gamma +2q\cdot t_i)$, and setting $q=\{0.0,0.2,0.4,0.6,0.8,1.0\}$, $n=100$, $\tau=0.1$, and $R=0.1$. Finally, the \emph{travelling repairman} set has long time windows whereas the processing times are short, by multiplying $t_T$ with a constant $c=\{1.0,1.2,1.4,1.6,1.8,2.0\}$, and $n=100$, $\tau=0.1$, and $R=0.1$. Since multiplying $t_T$ by $c$ makes the instance less congested, we generate $n\cdot c$ orders to the keep the congestion ratio at a similar level. For each set we generate ten instances with the same parameters. Therefore we have $5\cdot10+6\cdot10+6\cdot10=170$ new instances in total.

Besides Sparrow,
we include two other algorithms which correspond to the state of the art:
ILS by Silva et al.~\cite{silva2018exact} and HSSGA by Chaurasia and Singh~\cite{chaurasia2017hybrid}. As obtaining the original implementations of these algorithms proved to be infeasible, we reimplemented the algorithms according to their descriptions with varying degrees of success. For our implementation of ILS the gaps are too dissimilar to the gaps reported for the benchmark instances provided by Cesaret et al.~\cite{cesaret2012tabu} to argue reproduction of their algorithm. Therefore we refer to our implementation as ILS*. For HSSGA using proposed parameters our gap was found to be at most 1\% larger than the values reported by Chaurasia and Singh~\cite{chaurasia2017hybrid}, with a similar count of instances solved to optimality for the same set of benchmark instances. We argue that in the latter case the implementation is at least globally representative of the performance of HSSGA as described in the article.

All the three algorithms are run on Intel Core i5 3.20GHz CPU with 8GB memory, using a single core. Sparrow uses the same parameters as mentioned in Section~\ref{sec_comparison} and the other two algorithms use the same parameters as mentioned in articles \cite{chaurasia2017hybrid,silva2018exact}.

\begin{figure}[htbp]
\centering
\includegraphics[width=2.3in]{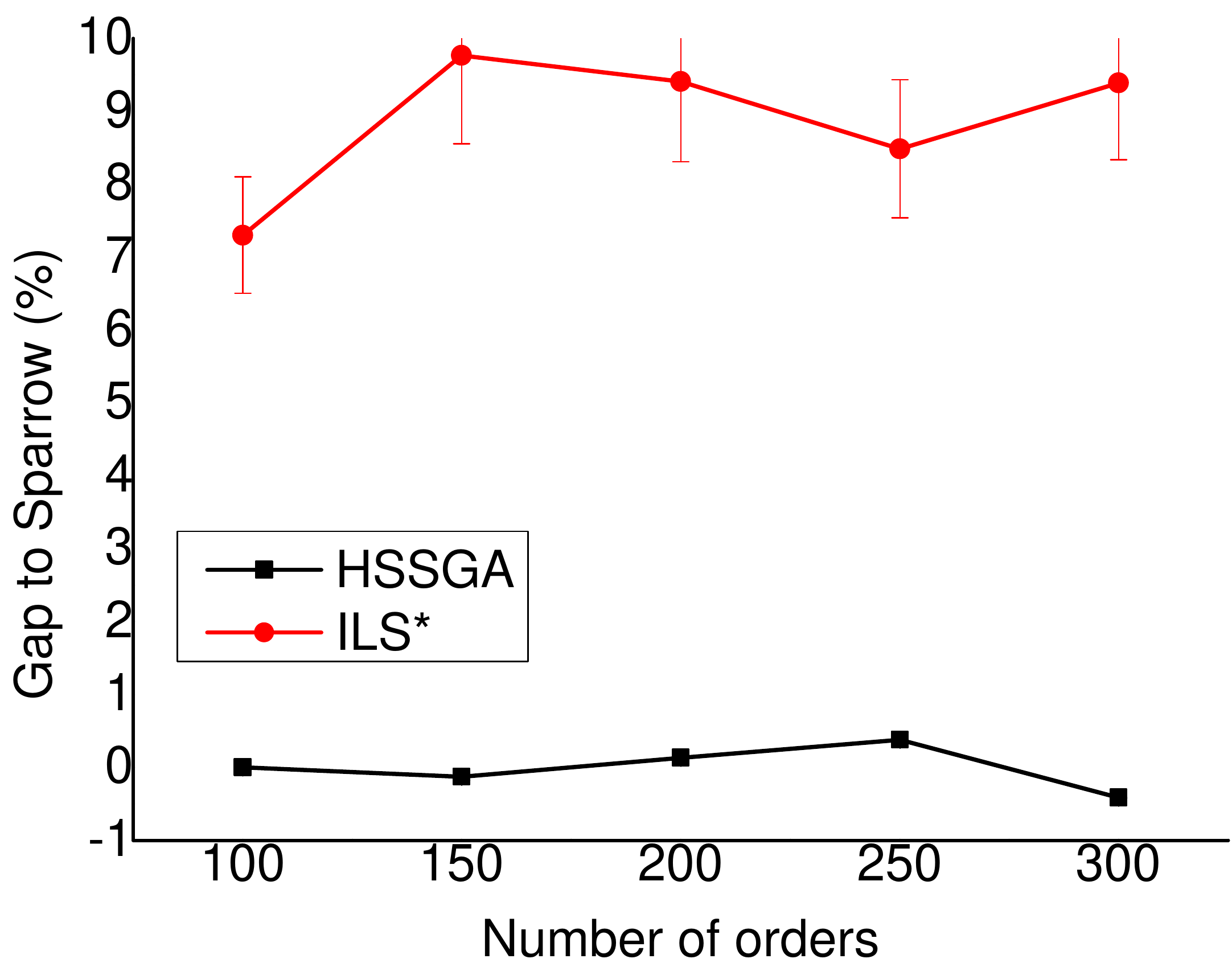}\hfill \includegraphics[width=2.3in]{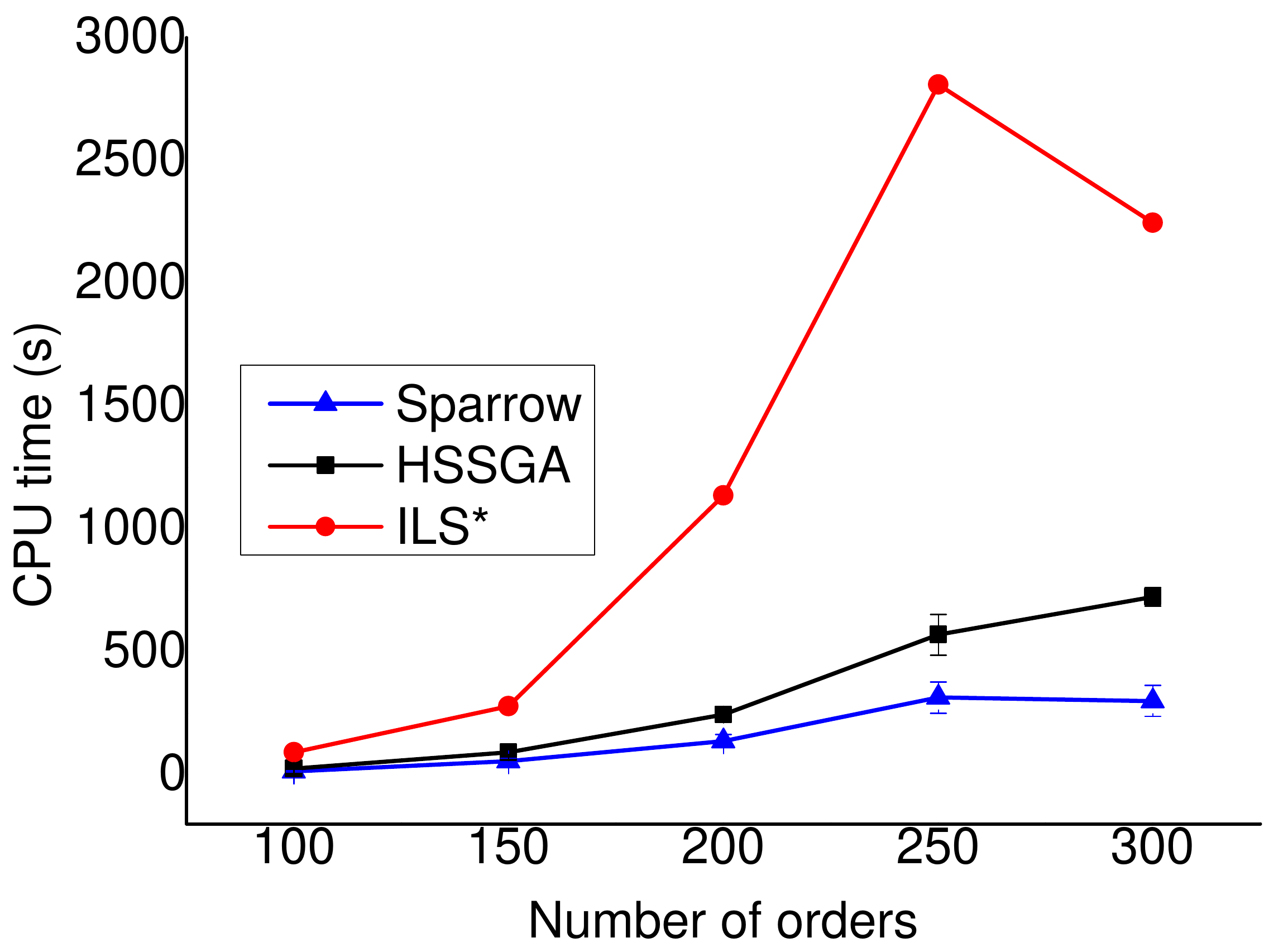}
\caption{The gaps of HSSGA and ILS* to Sparrow (Left), and the CPU times of three algorithms on the satellite scheduling set (right).}\label{figsatellite}
\end{figure}

\begin{figure}[htbp]
\centering
\includegraphics[width=2.3in]{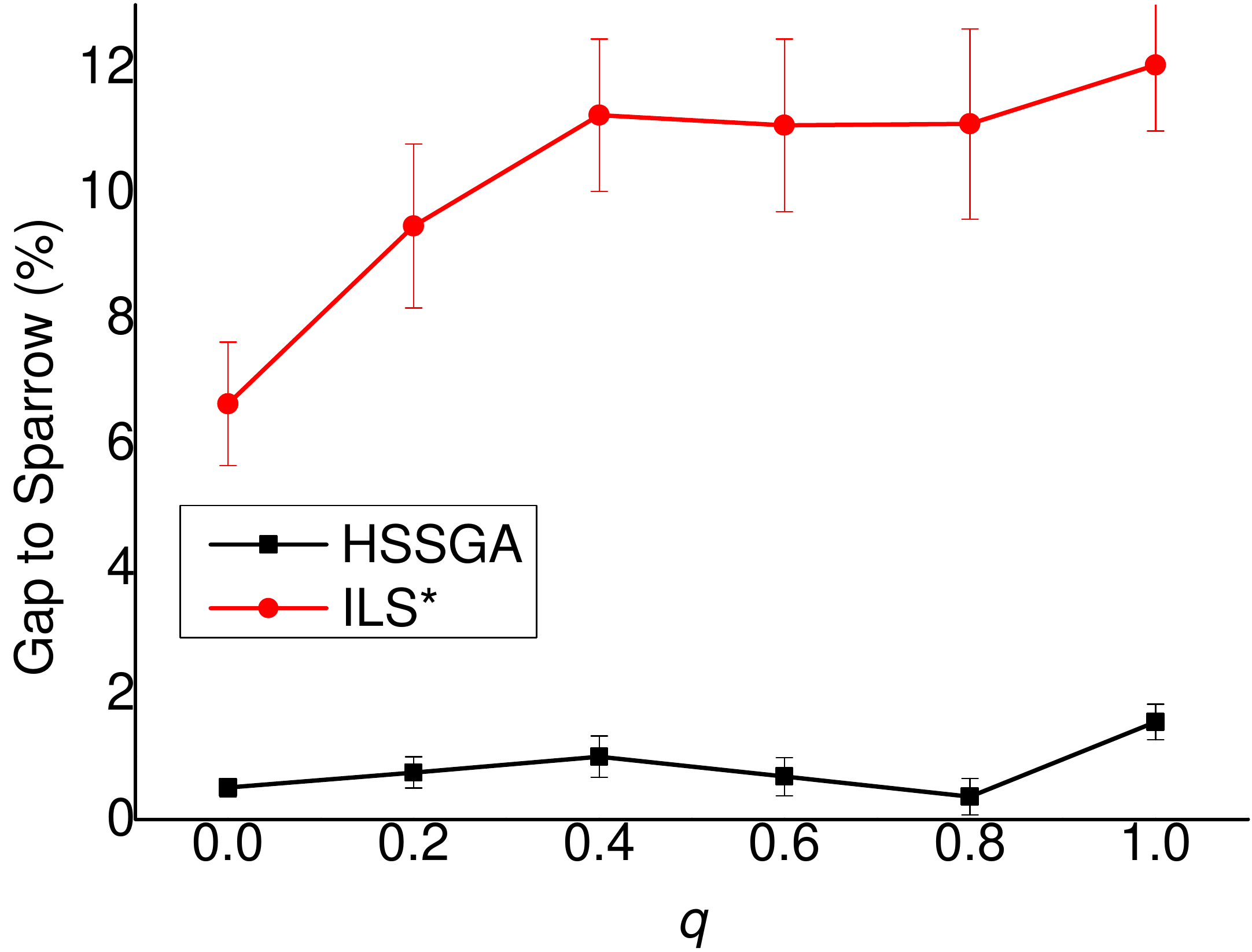}\hfill \includegraphics[width=2.3in]{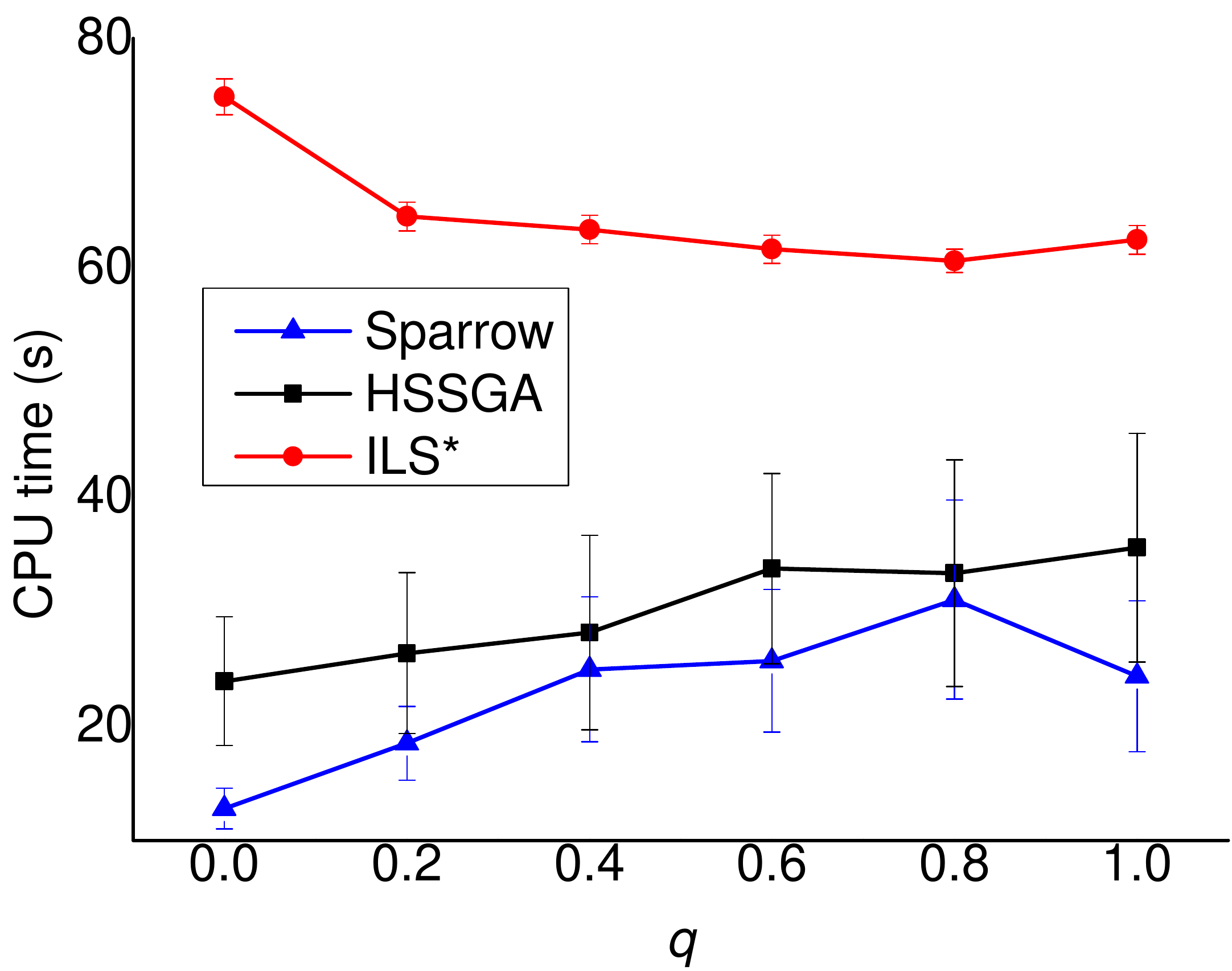}
\caption{The gaps of HSSGA and ILS* to Sparrow (Left), and the CPU times of three algorithms on the commerce set (right).}\label{figcommerce}
\end{figure}

\begin{figure}[htbp]
\centering
\includegraphics[width=2.3in]{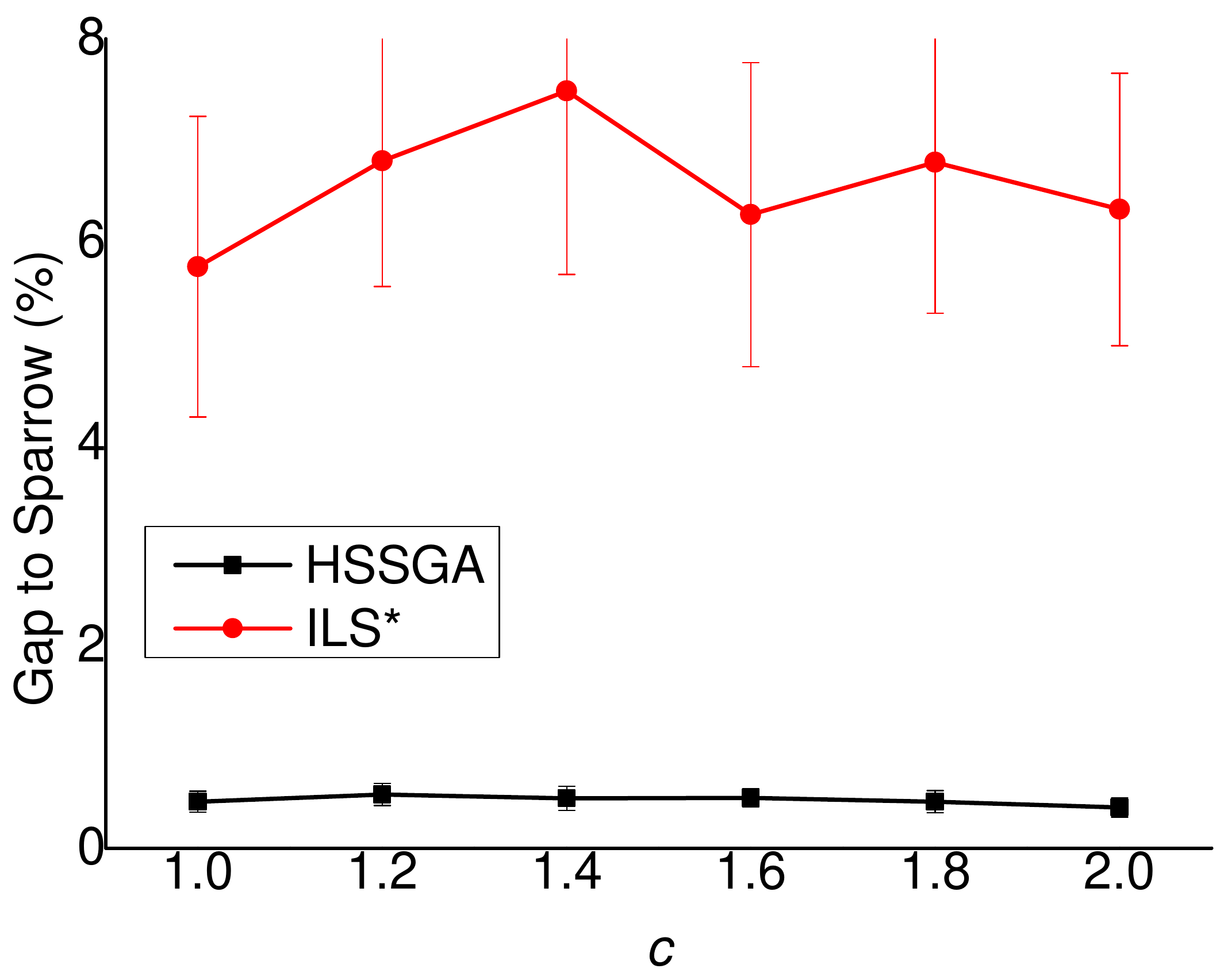}\hfill \includegraphics[width=2.3in]{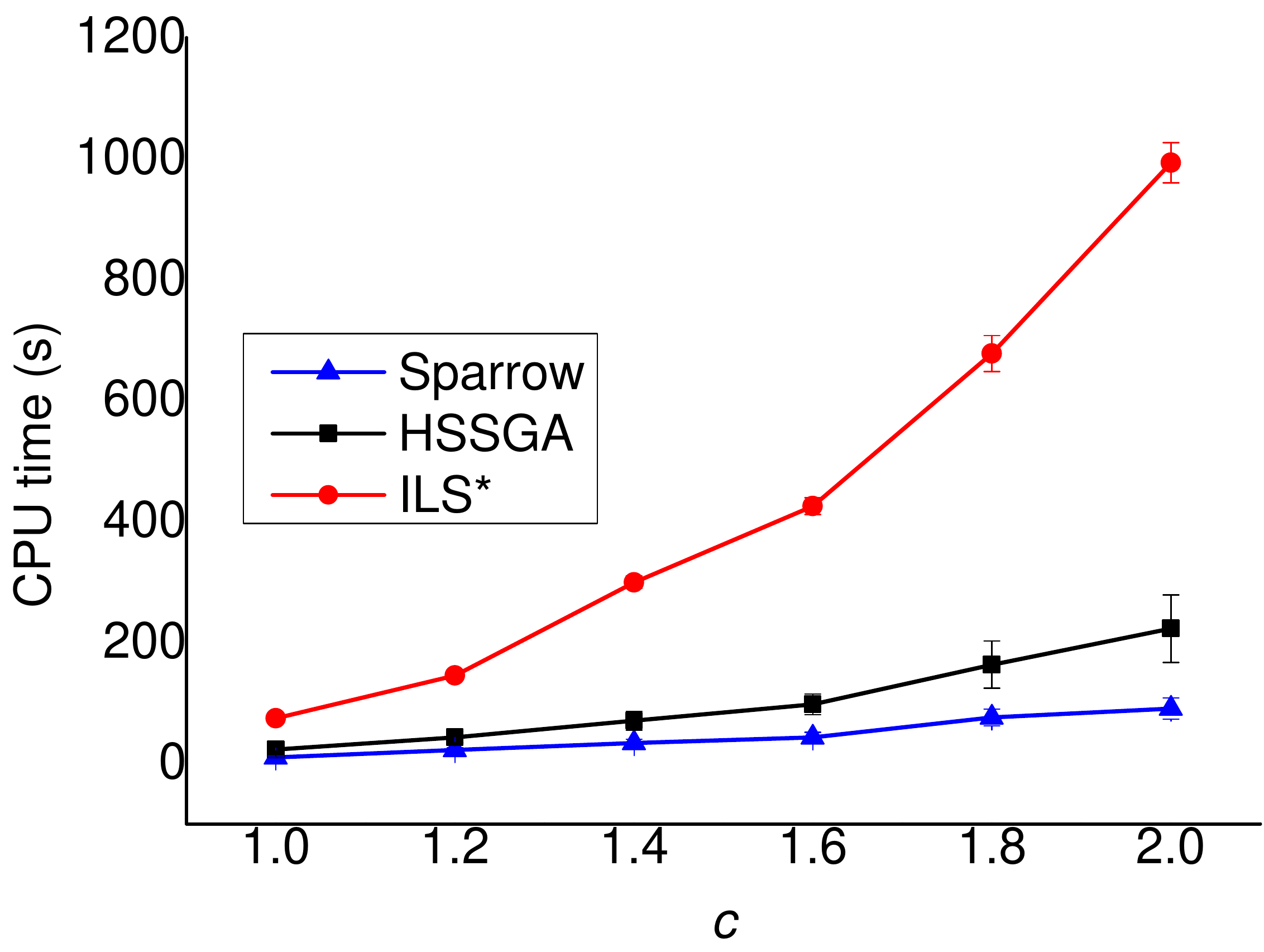}
\caption{The gaps of HSSGA and ILS* to Sparrow (Left), and the CPU times of three algorithms on the travelling repairman set (right).}\label{figtravelrepair}
\end{figure}

We do not have the upper bounds of the new instances. Therefore we calculate the gaps between the results of HSSGA and ILS* and the results of Sparrow. The gaps of HSSGA and ILS*, as well as the CPU times of the three algorithms on the three new instance sets are shown in Figures~\ref{figsatellite}-\ref{figtravelrepair}. Each point is the average result of 10 runs. Since the runtime of ILS* turned out to be too long for large instances, we set a time limit of 3600s.

On the satellite scheduling set, the gap between Sparrow and ILS* increases when there are more orders, which shows that ILS* cannot handle larger instances well. The gap between Sparrow and HSSGA does not change much. This shows that both Sparrow and HSSGA have a stronger ability to solve instances with larger congestion ratio efficiently. For instances with 150 and 300 orders, HSSGA performs better than Sparrow (with the gap smaller than 0). Regarding CPU times, ILS* shows a weak scalability. There is decline at order size 300, because many instances in order size 300 were stopped by the time limit, resulting in a lower average CPU time. Sparrow uses the least time. Both HSSGA and Sparrow algorithm are based on a hybridization of a population-based method and a local search method. The two algorithms integrate the exploration ability of the population-based method and the exploitation ability of the local search method well, making them adapt well for instances with varying sizes.

On the commerce set, the gaps of ILS* and HSSGA both increases when $q$ is larger (i.e., when revenue and processing time are more correlated), which shows that Sparrow can solve instances with correlated processing time and revenue well. Sparrow uses the least time and produces the best solutions. Sparrow performs well on this set of instance mainly because of the self-adaption ability of the ALNS component: ALNS uses multiple neighbourhood operators and can choose the most efficient one according to the instances. When the processing time and revenue of orders are more correlated, the operators based on the unit revenue would become less efficient to select orders. In this case Sparrow chooses other operators and the correlation between processing time and revenue does not influence the performance of Sparrow much.

Last, on the travelling repairman set, the changes of the gaps of the two algorithms are not clear. However it is obvious that Sparrow uses the least time and produces the best solutions. We believe this is because of the fast insertion algorithm with the time slack strategy, which provides a good flexibility for the orders to shift in the long time windows.

As a summary, Sparrow has a relatively high ability to handle harder and real-life cases, especially when revenue and processing time are more correlated, and the time window is very long. The HSSGA algorithm performs well on large
instances in terms of optimality gap, although taking more time than
Sparrow. The ILS* algorithm performs worst on these harder instances, and has the worst scalability.

\section{Conclusions}\label{sec_con}

This article studied the Order Acceptance and Scheduling (OAS) problem
with sequence-dependent setup times and time windows.  We proposed a
novel memetic algorithm called Sparrow, a hybridization of
the biased random key genetic algorithm (BRKGA) and adaptive large
neighbourhood search (ALNS).  We introduced a new bounded-width gene
encoding strategy, a hybrid decoding method, an intelligent crossover
operator, several neighbourhood operators, and a fast insertion
algorithm, making Sparrow suitable for problems with
varying properties.

Sparrow is tested on a set of standard benchmark with 750 instances with 25--100 orders. Compared with state-of-the-art algorithms, Sparrow obtains better-quality solutions with comparable running time. The gaps between Sparrow and other algorithms tend to increase when the problem instances get more difficult.  We further study this problem by analyzing the correlation between problem properties and the algorithm performance and find that the congestion ratio of the instance, the length of time windows and the correlation between revenue and processing time are the key properties influencing the difficulty of the problem.  Finally, we generate new instances with up to 300 orders, longer time windows and more correlated processing time and revenue, and compare the performances of different algorithms on these new realistic instances to understand their strengths and weaknesses. We find that Sparrow has a good ability to solve these difficult instances and the HSSGA algorithm from \cite{chaurasia2017hybrid} performs well on large instances in terms of optimality gap, although taking more time than
Sparrow. We believe that the integration of the exploration ability of population-based methods and the exploitation ability of local search methods in Sparrow and HSSGA, and the varying neighbourhood operators and the fast insertion strategy in Sparrow make them successful on the difficult instances.

Our next steps are to further improve the performance of Sparrow on large instances, and apply it to other real-world problem domains such as the agile satellite observation scheduling problem which has time-dependent revenue, time-dependent setup times and multiple machines and time windows.

\section*{Acknowledgements}

We gratefully thank Dr.~Yuri Laio T.V. Silva and Dr.~Anand Subramanian for providing the tight upper bound of each instance. This work is supported by China Scholarship Council for Lei He's visit at Delft University of Technology, the Netherlands (Grant No.~201703170269), and China Hunan Postgraduate Research Innovating Project (Grant No.~CX2018B020).

\section*{References}

\bibliography{caie2019paper}

\end{document}